\documentclass[lettersize,journal]{IEEEtran}

\usepackage{array}
\usepackage{stfloats}
\usepackage{url}
\usepackage{amsmath,amssymb,amsfonts}
\usepackage{amsthm}
\newtheorem{myDef}{Definition}
\newtheorem{myTheo}{Theorem}
\usepackage{cite}
\usepackage{pifont}
\usepackage{algorithmic}
\usepackage[ruled,boxed,linesnumbered]{algorithm2e}
\usepackage{subfigure}
\usepackage{graphicx}
\usepackage{textcomp}
\usepackage{xcolor}
\usepackage{multirow}
\usepackage{bm}
\usepackage{makecell}
\usepackage{threeparttable}
\usepackage{comment}
\SetKwInput{KwInit}{Initialization}

\begin{document}

\title{MetaTrading: An Immersion-Aware Model Trading Framework for Vehicular Metaverse Services}

\author{Hongjia~Wu, Hui~Zeng,
	Zehui~Xiong, Jiawen~Kang, Zhiping~Cai,\\Tse-Tin Chan,~\IEEEmembership{Member,~IEEE}, Dusit~Niyato,~\IEEEmembership{Fellow,~IEEE}, and Zhu~Han,~\IEEEmembership{Fellow,~IEEE} 
        
\thanks{This work was supported in part by NSF under Grant \mbox{ECCS-2302469} and Grant \mbox{CMMI-2222810}; in part by Toyota; in part by Amazon; in part by the Japan Science and Technology Agency (JST) Adopting Sustainable Partnerships for Innovative Research Ecosystem (ASPIRE) under Grant \mbox{JPMJAP2326}; in part by NSFC under Grant \mbox{62472434}; in part by the Science and Technology Innovation Program of Hunan Province under Grant \mbox{2022RC3061}; and in part by the Research Matching Grant Scheme from the Research Grants Council of Hong Kong. \emph{(Corresponding author: Tse-Tin Chan.)}}
\thanks{H.~Wu and T.-T.~Chan are with the Department of Mathematics and Information Technology, The Education University of Hong Kong, Hong Kong, China (e-mail: {whongjia@eduhk.hk}; {tsetinchan@eduhk.hk}).}
\thanks{Z. Xiong is with the Pillar of Information Systems Technology and
Design, Singapore University of Technology and Design, Singapore (e-mail: zehui xiong@sutd.edu.sg).} 
\thanks{J. Kang is with the School
of Automation, Guangdong University of Technology, China (e-mail:
kavinkang@gdut.edu.cn).}
\thanks{H. Zeng and Z. Cai are with the College of Computer Science and Technology, National University of Defense Technology, China (e-mail: zenghui116@nudt.edu.cn; zpcai@nudt.edu.cn).}
\thanks{D. Niyato is with the College of Computing and Data Science, Nanyang Technological University, Singapore (e-mail: dniyato@ntu.edu.sg).}
\thanks{Z. Han is with the Department of Electrical and Computer Engineering at the University of Houston, Houston, TX 77004 USA, and also with the Department of Computer Science and Engineering, Kyung Hee University, Seoul, South Korea, 446-701 (e-mail: hanzhu22@gmail.com).}
}

\maketitle

\begin{abstract}
Timely updating of Internet of Things (IoT) data is crucial for achieving immersion in vehicular metaverse services. However, challenges such as latency caused by massive data transmissions, privacy risks associated with user data, and computational burdens on metaverse service providers (MSPs) hinder the continuous collection of high-quality data.
To address these challenges, we propose an immersion-aware model trading framework that enables efficient and privacy-preserving data provisioning through federated learning (FL).
Specifically, we first develop a novel multi-dimensional evaluation metric for the immersion of models (IoM). The metric considers i) the freshness and accuracy of the local model, and ii) the amount and potential value of raw training data.
Building on the IoM, we design an incentive mechanism to encourage metaverse users (MUs) to participate in FL by providing local updates to MSPs under resource constraints. 
The trading interactions between MSPs and MUs are modeled as an equilibrium problem with equilibrium constraints (EPEC) to analyze and balance their costs and gains, where MSPs as leaders determine rewards, while MUs as followers optimize resource allocation. To ensure privacy and adapt to dynamic network conditions, we develop a distributed dynamic reward algorithm based on deep reinforcement learning, without acquiring any private information from MUs and other MSPs.
Experimental results show that the proposed framework outperforms state-of-the-art benchmarks, achieving improvements in IoM of $38.3\%$ and $37.2\%$, and reductions in training time to reach the target accuracy of $43.5\%$ and $49.8\%$, on average, for the MNIST and GTSRB datasets, respectively. These findings validate the effectiveness of our approach in incentivizing MUs to contribute high-value local models to MSPs, providing a flexible and adaptive scheme for data provisioning in vehicular metaverse services.

\end{abstract}


\section{Introduction}\label{I}
\IEEEPARstart{T}{he} metaverse, envisioned as a major evolutionary step for the Internet, aims to create a fully immersive, self-sustaining virtual environment for activities such as playing, working, and socializing~\cite{wang2022survey}. This vision is propelled by advances in fifth- and sixth-generation (5G/6G) communication technologies, which offer low latency and high data throughput. These technologies play a critical role in seamlessly integrating the Internet of Things (IoT) data into metaverse services, thus bringing the once-fictional concept of immersive experiences closer to reality.
\begin{figure}[t]
	\centering
	\includegraphics[width=\linewidth]{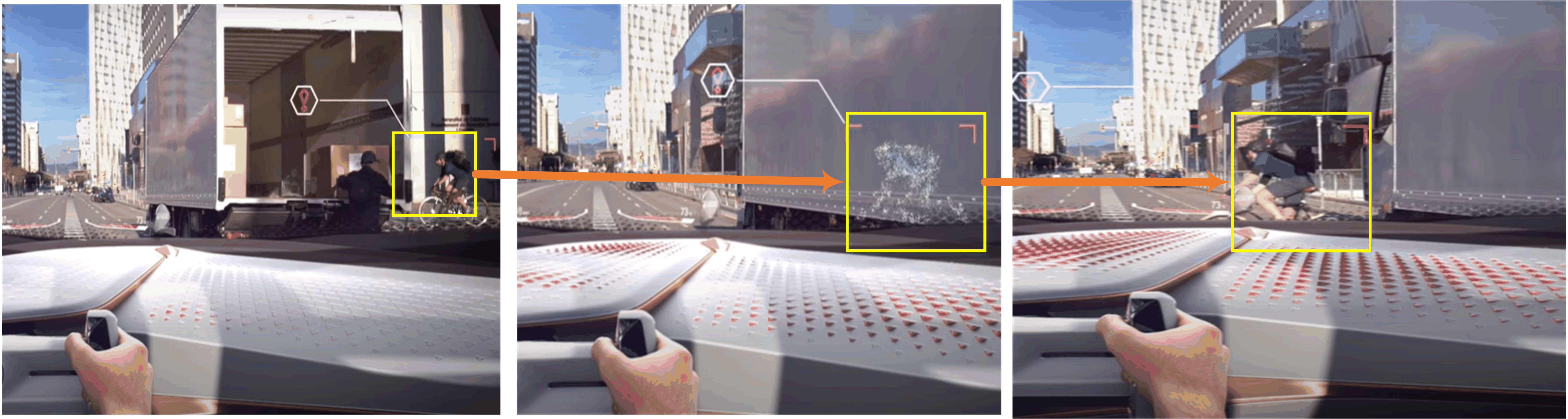}
	\caption{An example of AR services in the vehicular metaverse\protect\footnotemark. }
	\label{scenario}
\vspace{-0.1in}
\end{figure}\footnotetext{Image source: \url{https://www.jasoren.com/ar-in-automotive/}}

Metaverse services are beginning to reveal their vast potential across a broad spectrum of industries, from gaming and autonomous driving to education and marketing. Notably, the application of vehicles within the metaverse has attracted significant interest, particularly for the enhanced traffic safety enabled by state-of-the-art augmented reality (AR) technologies.
Market report~\cite{marketsandmarkets2023metaverse} projected that the global automotive metaverse market would grow from \$$1.9$ billion in 2022 to \$$6.5$ billion by 2030. Automakers such as BMW are actively investing in AR technologies. As shown in Fig.~\ref{scenario}, augmented information improves driving safety by displaying potential hazards obscured by the vehicle ahead. Moreover, Nissan's~\cite{10401029} upcoming technology utilizes a 3D AR interface that merges the real world with the virtual world to provide the driver with augmented information about the surrounding area. In addition, Neuron mobility\footnote{\url{https://immersive-technology.com/augmentedreality/neuron-introduces-new-ar-parking-assistant/}} has announced the launch of an innovative AR parking assistance system to improve passengers' parking routines and trip-end experiences.       
\IEEEpubidadjcol

The ability to capture information from the ``real'' world, particularly the ability to collect and process massive amounts of data from IoT devices, is the key to determining the success of immersive services (e.g., AR) in the vehicular metaverse. Meanwhile, the data must be processed and presented in a meaningful and responsive manner with adequate privacy protection. Technically, a high-quality AR experience relies on accurately detecting and classifying real-world objects (e.g., cars and pedestrians) under complex conditions~\cite{10562335}. To achieve this goal, sufficient valid data need to be collected and deeply processed to detect and classify objects accurately. Therefore, it is essential to focus on effectively collecting, processing, and protecting the data that supports a safer and more enjoyable driving experience.

\textbf{Motivations.}
The widely used data collection method, as adopted by Nissan~\cite{nissan_i2v_2023v} and studied in~\cite{GOKASAR2023119192}, involves gathering massive amounts of data through vehicle sensors, cameras, and roadside devices, and all data being processed centrally. While this approach is effective for various applications, when it comes to the situation with multiple metaverse users (MUs) and metaverse service providers (MSPs) associated with different companies, the centralized data collection approach is not applicable and may lead to the following issues. First, AR services in the vehicular metaverse need to be highly immersive so that MUs feel fully immersed in the rendered environment, such as visualized driving. However, data synchronization is hindered by the latency from massive real-time data updates under unstable and resource-limited communication conditions. Note that the value of real-time data diminishes over time~\cite{10552896}. Also, delays can severely impact the MU's experience and cause dizziness~\cite{10772049}. Second, the data to be transmitted may be sensitive and private, such as location, movement, and biometrics, which can create a better immersive experience but may inevitably increase the privacy risk of MUs. 

Federated learning (FL) has been adopted in prior work~\cite{AR,10304077,10562335,10758814} to enable collaborative model training without sharing raw data (e.g., sensor/imaging data from vehicles). In this approach, local updates are uploaded by individual MUs to MSPs for AR services, rather than transmitting large volumes of data centrally, thus significantly reducing the communication burden. 
However, MUs are typically self-interested and are reluctant to share local models with MSPs due to the additional computation, communication, and energy overhead. To overcome this issue, incentive mechanisms using strategies such as contract theory~\cite{10254627}, Stackelberg game~\cite{10485381,li2025satisfaction} and multi-winner
sealed-bid auction~\cite{10.1145/3599971}, have been proposed to encourage MUs to contribute local models. However, existing studies fail to explicitly assess the model value from multiple dimensions, making it difficult for both MSPs and MUs to quantify the benefits of the models for MSPs. Furthermore, most of these solutions are designed for a single MSP and do not consider the joint optimization of MUs' limited computational and communication resources. As a result, an efficient and privacy-preserving framework for data synchronization is needed to improve the immersive AR experience in the vehicular metaverse. 

To address the above research gaps, we propose an immersion-aware model trading framework for a multi-MSP, multi-MU vehicular metaverse. This framework is designed to incentivize MUs to become active contributors by providing local models tailored to the specific needs of MSPs. However, MUs have different sampling costs and limited computational and communication resources, while MSPs differ in their model preferences and compete with each other. These inherent asymmetries and competitive dynamics make it challenging to establish an efficient model trading ecosystem. This leads to two key research questions that underpin our work:
\begin{itemize}
    \item {How can MUs be effectively incentivized to contribute high-value local models that benefit vehicular metaverse services for diverse MSPs?} 
    
    \item {How can the dynamic competitive interactions among MSPs be modeled to achieve a trade-off in gains between MUs and MSPs, ensuring sustainable and stable trading of local models in the vehicular metaverse?}   
\end{itemize}

\begin{figure}[t]
	\centering
	\includegraphics[width=\linewidth]{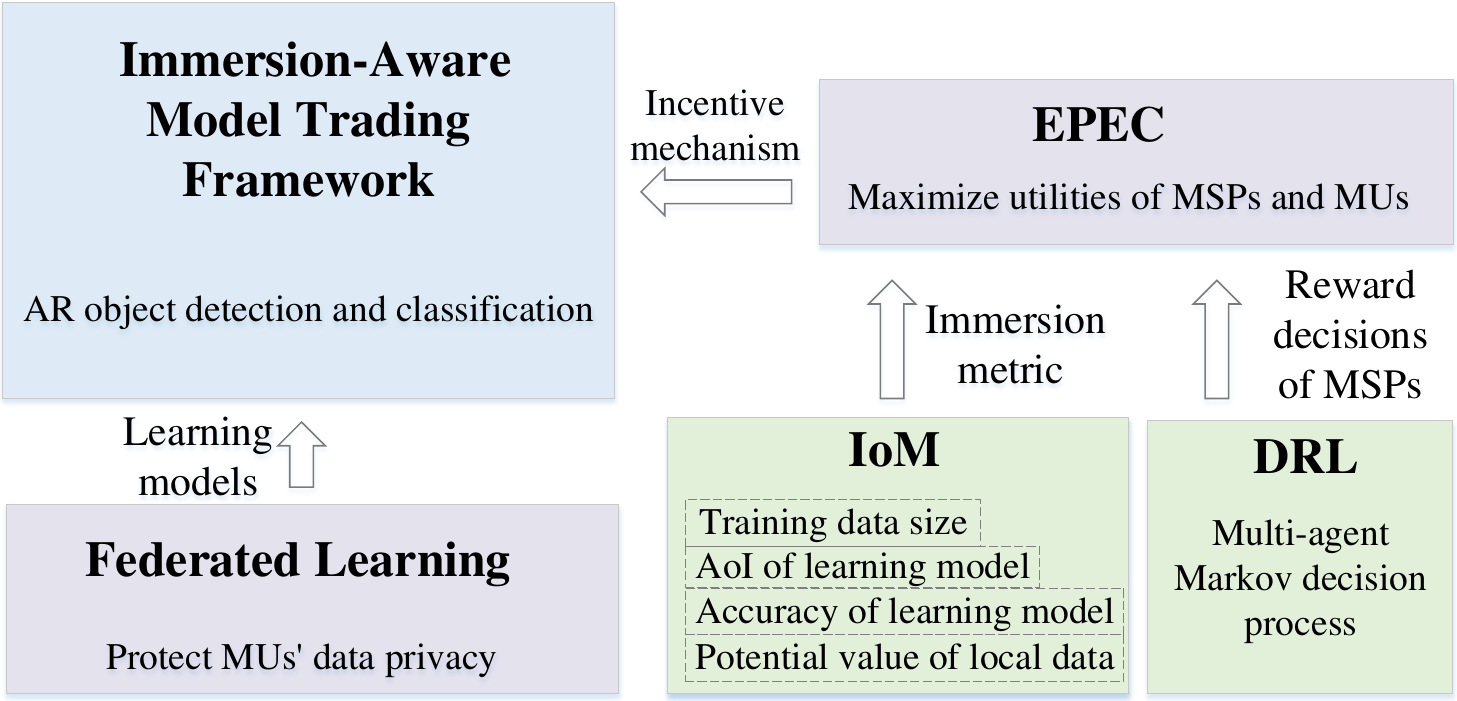}
	\caption{The outline of an immersion-aware framework for FL-assisted vehicular metaverse.}
	\label{tech}
\vspace{-0.1in}
\end{figure}

\textbf{Proposed Framework and Contributions.} As depicted in Fig.~\ref{tech}, our model trading framework consists of four components: FL mechanism, metric design, game modeling, and dynamic algorithms. Specifically, this work first designs a new metric called the \underline{i}mmersion \underline{o}f the learning \underline{m}odel (IoM) to evaluate the value of local models contributed by MUs to MSPs. 
This metric jointly considers the freshness and accuracy of the local model, as well as the amount and potential value of raw training data.
Building on this metric, we propose an immersion-aware incentive mechanism that aligns the interests of both MUs and MSPs.
Then, we model the dynamic competitive trading interactions as an equilibrium problem with equilibrium constraints (EPEC), which is a hierarchical optimization problem with equilibria at two levels~\cite{EPEC}. Moreover, given the dynamic networks and the privacy concerns of MSPs, we formulate the reward decisions of MSPs as a multi-agent Markov decision process (MAMDP) and develop a \underline{m}ulti-agent \underline{d}eep reinforcement learning (DRL)-based \underline{d}ynamic \underline{r}eward (MDDR) approach to obtain the reward decisions in a fully distributed manner. 
In summary, the key contributions of this work are as follows.
\begin{itemize}{
\item \textit{Incentive mechanism design for immersion-aware model trading.} From the perspectives of both MUs and MSPs, we propose an incentive mechanism that encourages MUs to contribute high-value local models tailored to the specific demands of MSPs. To our knowledge, this is the first study focusing on incentive mechanism design for efficient and privacy-preserving data synchronization in multi-MU and multi-MSP vehicular metaverse environments.

 \item \textit{Novel design of multi-dimensional metric.} To quantify immersion enhancement provided by MUs for AR services, we design an immersion metric of the local model integrating four critical dimensions: the freshness and accuracy of the local model, as well as the amount and potential value of raw training data. Freshness is captured through age of information (AoI), while potential value is evaluated by the difference between model predictions and true
labels. This metric enables fine-grained evaluation of model value under resource constraints and provides a basis for incentive and decision-making strategies in model trading.

  \item \textit{Theoretical analysis and algorithm designs.} 
Given multiple resource-constrained MUs and competing MSPs, we model their interactions as an equilibrium problem with equilibrium constraints (EPEC). We theoretically prove the existence and uniqueness of the equilibrium, and further develop a fully distributed MDDR approach that adapts to dynamic environments and operates without accessing any private information of MUs or MSPs.
  
     \item \textit{Performance evaluation.} Extensive numerical simulations are conducted based on AR-related vehicle datasets to validate the efficacy and efficiency of MDDR and the proposed immersion-aware model trading framework. Numerical results show that our proposed mechanism improves the IoM by $38.3\%$ and $37.2\%$, and reduces the training time to reach the target accuracy by $43.5\%$ and $49.8\%$, on average, for the MNIST and GTSRB datasets, respectively, compared with state-of-the-art benchmarks.}
\end{itemize}

The rest of this paper is organized as follows.
Section~\ref{II} discusses the related work. In Section~\ref{III}, we present
the system overview and design the immersion metric of the local model. Section~\ref{IV} gives the game formulation, and Section~\ref{V} analyzes the existence of the equilibria at two levels. In Section~\ref{VI}, we give the detailed design of MDDR. Section~\ref{VII} shows numerical experiments to evaluate the framework performance, and finally,  Section~\ref{VIII} concludes the paper.

\section{Related Work}\label{II}
In this section, we discuss the work related to our study in terms
of edge-enabled vehicular metaverse services, FL for AR, and incentive mechanisms for data synchronization in the metaverse.

\subsection{Edge-Enabled Vehicular Metaverse Services}

Vehicular metaverse services require ultra-low latency and high reliability to enable immersive experiences, such as AR navigation and real-time digital twins. To meet these demands, researchers have extensively explored edge-based strategies for task offloading and resource management. For instance, Feng \textit{et al.}~\cite{Feng} proposed an innovative resource allocation framework for AR-empowered vehicular edge metaverse, which significantly improves data utility and reduces vehicle energy consumption through multi-dimensional optimization. Similarly, Khan~\textit{et al.}~\cite{Khan_2024} introduced a cooperative framework integrating task offloading, sensing, learning, and communication to reduce both transmission energy consumption and latency in resource-constrained environments. To cope with dynamic network conditions, Tong~\textit{et al.}~\cite{10608164} proposed an attribute-aware auction mechanism for resource allocation, and applied a GPT-based deep reinforcement learning algorithm to adjust the auction clocks.
Regarding security and privacy, Kang~\textit{et al.}~\cite{Kang} presented a cross-metaverse empowered dual pseudonym management framework to mitigate privacy leakage risks during the dynamic communications among vehicular edge metaverses. Moreover, a novel four-layer security framework was designed to protect vehicular metaverse systems from data poisoning attacks~\cite{10892270}.

In addition to theoretical models and optimization frameworks for integrating autonomous vehicles into the metaverse, several studies have proposed practical decision-making approaches to evaluate and select among alternative implementation strategies. For example, Deveci~\textit{et al.}~\cite{DEVECI2023122681} proposed a framework that combines a fuzzy logarithmic weighting method with a similarity-based ranking technique to assess four traffic safety solutions enabled by the metaverse. Their results show that public transportation is the most practical and scalable option. Moreover, they applied a decision-making method based on Q-rung orthopair fuzzy sets to rank personal mobility alternatives, offering practical guidance for system planning in~\cite{9827997}. Gokasar~\textit{et al.}~\cite{GOKASAR2023119192} introduced a two-stage fuzzy decision-making framework to evaluate integration options for self-powered sensors in autonomous vehicles. The study concludes that using these sensors for real-time traffic management in the metaverse is the most effective solution.

These works, from diverse perspectives, have significantly advanced the security, reliability, and performance of vehicular metaverse services, providing valuable insights for future research. Notably, distinct from existing studies, our work focuses on designing an incentive mechanism that transforms MUs into active contributors. These contributors efficiently provide high-value local models to MSPs, thereby enhancing the overall service experience.

\subsection{Federated Learning for AR in the Metaverse}
Privacy preservation is critical in vehicular metaverse services, particularly when training models using sensitive user data. Federated learning (FL) has emerged as a promising technique to enable collaborative model training without sharing raw data. For example, Chen~\textit{et al.}~\cite{AR} proposed a framework that integrates mobile edge computing and FL to solve the computational efficiency and low-latency object classification problem for AR services. However, they did not address how communication and computational resources should be allocated within the FL framework. 

To overcome this limitation, Zhou~\textit{et al.}~\cite{10304077} developed an FL-assisted MAR system via non-orthogonal multiple access for the metaverse. Furthermore, a resource allocation algorithm was designed to balance accuracy and energy consumption.
To further improve learning efficiency, an adaptive and resource-efficient FL algorithm tailored for AR applications was proposed in~\cite{10562335}. This approach mitigates the effects of non-IID data distributions, reduces resource usage, and enhances the quality of experience.
Additionally, Hazarika~\textit{et al.}~\cite{10758814} explored the integration of quantum computing with FL to provide a cost-efficient and adaptive solution for the dynamic nature of vehicular environments.
Nadimi~\textit{et al.}~\cite{nadimi2025} proposed a multi-modal, multi-task federated foundation model (FedFMs) architecture. It combines the privacy-preserving benefits of FL with the representation capabilities of foundation models to address sensor diversity, personalized interactions, and device heterogeneity in extended reality systems.

Incorporating FL into AR services brings the benefits of reduced communication latency and privacy protection.
However, the existing studies fail to consider the integration of FL into AR for the multi-MSP and multi-MU metaverse scenario, while overlooking the selfishness of MUs, i.e., whether they are willing to participate in FL learning.

\begin{table}[t]
\caption{Incentive Mechanisms for Data Synchronization}
\label{related}
\begin{tabular}{c|c|c|c|c}
\hline
Ref.                                                & \begin{tabular}[c]{@{}c@{}}MSPs\\ MUs\end{tabular} & \begin{tabular}[c]{@{}c@{}} Theory \\ Method\end{tabular}                                                                  &  \begin{tabular}[c]{@{}c@{}}  Main \\ Metric    \end{tabular}                                                                       & \begin{tabular}[c]{@{}c@{}}Data \\Privacy \end{tabular} \\ \hline
{\cite{10254627}}                                             & $\times$                                  &    \begin{tabular}[c]{@{}c@{}} Contract theory \\under PT     \end{tabular}                                                               & AoI                                                                         & \checkmark                                             \\ \hline
{\cite{10485381}}                                            & $\times$                                                  & \begin{tabular}[c]{@{}c@{}}Dual \\ game-theoretic \end{tabular}                                                                   & \begin{tabular}[c]{@{}c@{}}Reputation \\ scores \end{tabular}  &  \checkmark                                               \\ \hline
{\cite{li2025satisfaction}}                                            & $\times$                                 & Stackelberg game                                                                   & \begin{tabular}[c]{@{}c@{}}Model quality \\ and latency  \end{tabular}                                                           & \checkmark                                              \\ \hline
{\cite{10.1145/3599971}}                                            & \checkmark                                                   & IISG and PMS-AM                                                                    & \begin{tabular}[c]{@{}c@{}}Model quality \\ and freshness  \end{tabular}                                                                            & \checkmark                                         \\ \hline
{\cite{10810728}}                                            & \checkmark                                                   & \begin{tabular}[c]{@{}c@{}} Coalition formation, \\Stackelberg game      \end{tabular}                                                                  & \begin{tabular}[c]{@{}c@{}} Volume of \\sensing data      \end{tabular}                                                                      & $\times$                                             \\ \hline
{\cite{10623271}}                                            & \checkmark                                 &  \begin{tabular}[c]{@{}c@{}} Deep learning \\based auction \end{tabular}                                                                      & \begin{tabular}[c]{@{}c@{}} Quality,\\reliability, latency\end{tabular}       & $\times$                                             \\ \hline
{\cite{9583902}}                                            & \checkmark                                                  & Contract theory                                                              & Computing delay                                                                            & $\times$                                             \\ \hline
{\cite{ZHAO2025101507}}                                            & \checkmark                                                  & Stackelberg game  & \begin{tabular}[c]{@{}c@{}}Energy \\ consumption    \end{tabular}                                                             & $\times$                                             \\ \hline
{\cite{10807929}}                                            & \checkmark                                 & \begin{tabular}[c]{@{}c@{}}Multi-objective \\ optimization \end{tabular}         &\begin{tabular}[c]{@{}c@{}} Peak signal-\\to-noise ratio \end{tabular}      & \checkmark                                              \\ \hline
\begin{tabular}[c]{@{}c@{}}Our\\  work\end{tabular} & \checkmark                                 & EPEC               & IoM                                                                              & \checkmark                            \\ \hline
\end{tabular}
\end{table}

 \subsection{Incentive Mechanism for Data Synchronization in the Metaverse.}

Immersive metaverse services depend on high-quality real-world sensing data. However, the process of collecting such data is often costly and may raise privacy concerns. To this end, various incentive mechanisms have been proposed to encourage MUs to contribute data and participate in model training.
 For instance, Zhang~\textit{et al.}~\cite{10810728} introduced a vehicle-assisted data sensing framework that incentivizes vehicles to upload sensing data, enhancing the driving experience.
Xu~\textit{et al.}~\cite{10623271} developed a UAV swarm system using digital twins and semantic communication to improve data synchronization and reduce latency. 
Lin~\textit{et al.}~\cite{9583902} designed an incentive-based congestion control scheme for digital twin edge networks, which addresses the stochastic service demand and the long-term provider profit by combining Lyapunov optimization and contract theory. Zhao~\textit{et al.}~\cite{ZHAO2025101507} proposed a physical reality incentive mechanism based on a two-stage Stackelberg game to motivate MUs to collect physical reality sensing data for the metaverse. 
Although these studies successfully encourage raw data contributions, they also pose a risk to user privacy due to potential data leakage. 

To address privacy concerns, Kang~\textit{et al.}~\cite{10254627} developed a cross-chain FL framework with an AoI-based contract theory model under prospect theory (PT) to incentivize data sharing. They utilized a hierarchical cross-chain architecture with a main chain and multiple subchains to perform decentralized, privacy-preserving, and secure data
training in both virtual and physical spaces. Baccour~\textit{et al.}~\cite{10485381} designed a dual game-theoretic framework for federated meta-learning (FML), incorporating a reputation system and Stackelberg game to enhance privacy, reduce energy consumption, and improve model performance. Li~\textit{et al.}~\cite{li2025satisfaction} designed a new metric named ``Satisfaction" to balance training latency and model quality. They integrated this metric into utility functions to incentivize node participation in FL while optimizing resource allocation.
However, these works are designed for a single MSP scenario, limiting scalability and applicability in multi-server scenarios commonly encountered in large-scale metaverse environments. 

To fill this gap, Zhang~\textit{et al.}~\cite{10.1145/3599971} employed an imperfect information Stackelberg game (IISG) to optimize the strategies of MUs and MSPs, and introduced a privacy-preserving multi-winner sealed-bid auction mechanism (PMS-AM) to enhance the quality of non-fungible token (NFT) with FL assistance. Esmail~\textit{et al.}~\cite{10807929} proposed a metaverse FL and semantic communication (MFSC) framework that matches IoT device capabilities to training tasks, optimizing digital twin quality through adaptive resource allocation.
Nonetheless, existing incentive criteria are often simplistic and fail to evaluate the values of contributed models comprehensively.
 
Unlike these prior works, our study focuses on vehicular metaverse services in multi-MSP and multi-MU scenarios. We propose a novel metric, IoM, which evaluates model value from multiple dimensions. Based on this metric, we design an incentive mechanism that better motivates MUs to contribute high-value local models, thereby improving the quality and user experience of metaverse services. A detailed comparison of
these approaches is provided in Table~\ref{related}, highlighting the key
contributions of this study.

\begin{figure}[t]
	\centering
	\includegraphics[width=\linewidth]{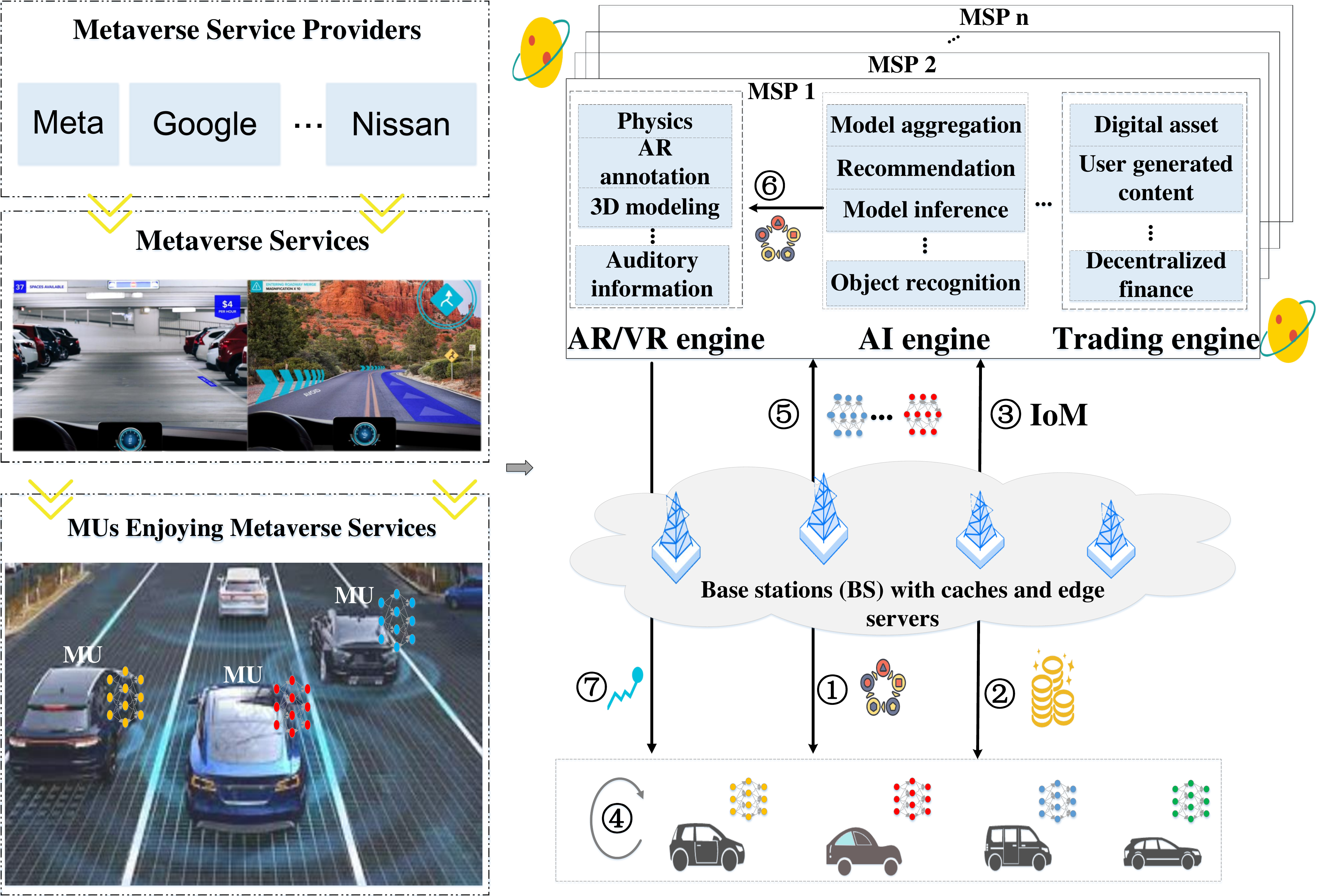}
	\caption{Workflow of the immersion-aware model trading framework\protect\footnotemark.}
	\label{overview}
 \vspace{-0.1in}
\end{figure}\footnotetext{Image source: \url{https://www.vanarama.com/blog/cars/4-ways-augmented-reality-will-revolutionise-the-automotive-industry}.}
\section{System Model}\label{III}
We first present the system overview of the immersion-aware model trading framework in Section~\ref{III-A}. 
Next, we introduce the FL mechanism adopted by our framework in Section~\ref{FLMECH} and design the immersion metric of the local model in Section~\ref{IoM}. 

\subsection{System Overview}\label{III-A}
Fig.~\ref{overview} illustrates the workflow of our immersion-aware model trading framework, designed to enhance the immersive and interactive AR experiences of MUs. MUs can contribute local models to the AR services of MSPs for better immersive experiences. For example, services such as object detection can be improved by training on street and pedestrian images captured by vehicles' cameras. In this way, AR services can be applied more effectively to visible driving, with more accurate and timely hazard warnings from windscreens.
The framework mainly consists of MSPs owned by different companies (e.g., Meta, Nissan, and Google), an infrastructure layer, and an interaction layer with the MUs that enjoy vehicular metaverse services. MSPs are powered by various technologies supporting vehicular metaverse services, such as AI engines, AR/VR engines, and metaverse trading engines. The infrastructure layer with base stations (equipped with caches and edge servers) provides the basis of 5G/6G communication services for the interactions among MUs and MSPs. MUs play essential roles as both consumers and contributors to vehicular metaverse services.
\begin{table}[t]
	\centering
	\caption{Definitions of Notation}
	\label{parameters}
	\begin{tabular}{l|l}
		\hline
		Notation     & Definition                                                  \\ \hline
		$m,M$            & Index of an MU, Number of MUs                                              \\
		$n,N$            & Index of an MSP, Number of MSPs                                               \\
		$\overline{\Delta}_{mn}$            &  The average AoI of MU $m$'s model provided for MSP $n$           \\
		$V_{mn}$           & The IoM contributed by MU $m$ to MSP $n$                                                   \\
		$f_{mn}$       & {MU $m$'s computational resource used for MSP $n$   }   \\
		$B_{mn}$       & {MU $m$'s communication resource used for MSP $n$   }   \\
		$X_{mn}$       & MU $m$'s training data set for MSP $n$        
		\\
		$\omega_{mn}$     &  The potential value of MU $m$'s local data for MSP $n$                \\
		$\theta_{m}$     &   MU $m$'s local accuracy threshold              \\
		\hline
	\end{tabular}
\vspace{-0.1in}
\end{table}

The framework adopts a multi-engine architecture, and this paper focuses on two components: the trading engine and the AI engine of the MSP.
The trading engine is responsible for incentivizing MUs to participate in mutually beneficial collaborations with MSPs, providing them with a trading platform to make their decisions (i.e., computational and communication resource allocation of MUs and reward decisions of MSPs). Then, each MU deciding to contribute utilizes the FL mechanisms integrated into AI engines to provide local models for MSPs under the strategic guidance of the trading results.
The specific workflow of the framework involves the following two phases.

\textbf{Phase I} (Incentive process):
Initially, MSPs' AI engines broadcast their global models to all MUs in the vehicular metaverse\footnote{The FL global models owned by different MSPs we consider could be object detection or classification models for intelligent terrain mapping, intuitive road safety, visible driving, etc. Note that it is feasible to train on multiple tasks with the same local data. For instance, for the same image, intuitive road safety focuses on pedestrians and vehicles, while intelligent terrain mapping focuses on landmarks and routes.} (\ding{172} in Fig.~\ref{overview}). Then, the trading engines of MSPs determine their digital currency prices per IoM, i.e., rewards to MUs, and broadcast them to MUs (\ding{173}). 
Next, each MU determines its allocation of computational and communication resources for contributing local models to various MSPs based on the IoM, costs, and rewards given by MSPs (\ding{174}).
Based on the responses from MUs, the trading engines of MSPs adjust the digital currency prices to maximize their own utilities. Steps \ding{173} and \ding{174} iterate until an agreement is reached between MUs and MSPs.

\textbf{Phase II} (FL process): Each MU, guided by decisions obtained through tradings, allocates computational resources to local training based on different global models from MSPs and generates local models for different MSPs (\ding{175}).  
Then, MUs upload their local updates to the corresponding AI engines of MSPs by consuming communication resources (\ding{176}).    
Following this, AI engines aggregate local updates from all MUs to update their global models and transmit them to their AR/VR engines for utilization (\ding{177}). Finally, the AR/VR engines provide enhanced AR services to MUs (\ding{178}). 
Steps \ding{175}-\ding{178} are repeated until the guidance time $T$ ends. To ease the presentation, we summarize some important notations
 in Table \ref{parameters}.

\subsection{FL Framework Mechanism Adopted}\label{FLMECH}
\begin{figure}[t]
	\centering
	\includegraphics[width=\linewidth]{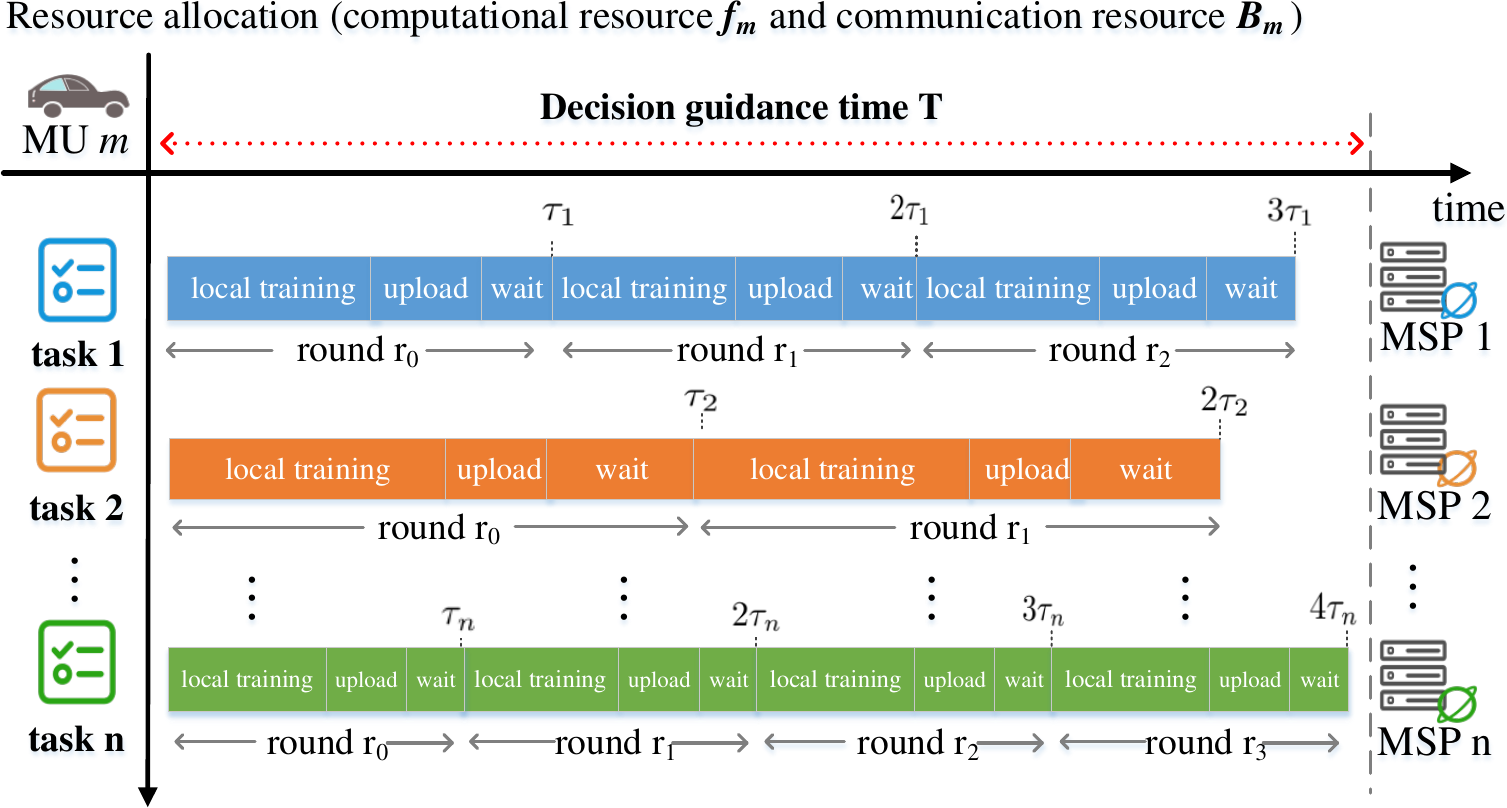}
	\caption{Illustration of the FL mechanism with $n$ tasks for MU $m$.}
	\label{fl_flow}
\vspace{-0.1in}
\end{figure}

We consider a set of $\mathcal{M} = \left \{ 1,\ldots,M \right \}$ MUs equipped with local computational capabilities and a set of $\mathcal{N} = \left \{ 1,\ldots,N \right \}$ MSPs with FL synchronous tasks\footnote{That is, the update time of the global model is limited by the slowest MU, i.e., the AI engine of the MSP must wait to receive local updates from all MUs before performing model aggregation. The advantages of synchronous FL are high model accuracy and fast convergence~\cite{liang}.}. Each MSP initiates an FL synchronous task with a virtual deadline $\tau_{n}\ (n \in \mathcal{N})$. The MUs can participate in FL to generate local models and contribute to enhancing vehicular metaverse services. 
As shown in Fig.~\ref{fl_flow}, MU $m$ provides local models to different MSPs simultaneously for a given time period $T$, meaning multiple FL tasks can be processed in parallel. Moreover, the data collected at the $(r-1)$-th round are used for local training at the $r$-th round.   

Without loss of generality, we consider that the trading decisions among MUs and MSPs are made at the beginning of period $T$.
Let \( X_{mn} = \{(y_i, z_i)\}_{i=1}^{|X_{mn}|} \) be the set of input-output pairs sampled from MU \( m \) for task \( n \), where \( y_i \) is the input with \( d \) features and \( z_i \) is its corresponding ground-truth label. Here, \( |X_{mn}| \) denotes the size of the set \( X_{mn} \). The data, such as images of streets and pedestrians, can be generated by high-definition cameras inside and outside the vehicles.
The steps involved in each iteration $r \in \left \{1,2,\ldots, R  \right \}$ are as follows: 
\begin{itemize}{
		\item {\textbf{\textit{Publishing Tasks}} :
			The AI engine of MSP $n$ broadcasts its global model $\jmath_{n}^{(r)}$ (e.g., object detection models used in AR services for visible driving and intuitive road safety) to all MUs in the $r$-th round.}
		
		\item {\textbf{\textit{Local Training}}: Each MU, for example, MU $m$, trains $\jmath _{n}^{(r)}$ with the most recently collected data $X_{mn}^{(r)}$ by stochastic gradient descent (SGD) within certain local rounds. The number of local training rounds depends on its local accuracy threshold.}
		
		\item {\textbf{\textit{Uploading and Aggregation}}: MUs transmit their local updates $\jmath _{mn}^{(r)}$ to the AI engine of MSP $n$. Upon receiving all the local updates, the MSP $n$'s AI engine aggregates them through the following weighted average:
			\begin{equation}
			\begin{aligned}
			\jmath_{n}^{(r+1)}&=\jmath ^{(r)}_{n}+\sum_{m=1}^{M}\frac{|X_{mn}^{(r)}|}{|X_{n}^{(r)}|}(\jmath_{mn}^{(r)}-\jmath_{n}^{(r)}),\\
			\end{aligned}
			\end{equation}
			and then obtains the new global model $\jmath_{n}^{(r+1)}$ for the next iteration.}
		
		\item {\textbf{\textit{Model Utilization}}: After global aggregation, MSP $n$'s AR engine obtains an updated global model that can be used for corresponding 3D modeling and other functions~\cite{AR}. The global rounds are iterated until a specific requirement is met, such as reaching a certain level of accuracy or a deadline.}}
\end{itemize}

\subsection{Immersion Metric of Local Model}\label{IoM}
MSPs determine the rewards for MUs based on the immersion of the local model (IoM) metric. Correspondingly, MUs need to decide the IoMs of local models provided to MSPs by allocating their computational and communication resources to maximize the benefits.
This process highlights the key role of ``IoM'' in the interactions between the two parties, serving as both an evaluation criterion and a basis for decision-making. Unlike traditional FL and incentive mechanisms, vehicular metaverse scenarios with immersion requirements necessitate consideration of both the contribution and freshness of the local model\footnote{For example, an MSP may receive fresher local updates but with lower contributions, or less fresh local updates but with higher contributions.
}, which will collectively affect the immersive experience of AR services, i.e.,  whether virtual objects can be placed in the physical world accurately and promptly.

Driven by the above considerations, we design the IoM based on four key dimensions: the freshness and accuracy of the local model, as well as the amount and potential value of raw training data. The freshness is captured through \textit{age of information} (AoI) $\Delta_{mn}$~\cite{10417012}, while the other three dimensions are jointly reflected in \textit{contribution prediction} $I_{mn}$. To measure the value of a local model that MU $m$ brought to the vehicular metaverse service of MSP $n$, we define the IoM as
\begin{equation}\label{iom}
V_{mn}=I_{mn}(\tau_{n}-\overline \Delta_{mn}),
\end{equation}
where the components are detailed as follows. In the rest of this paper, we refer to the trading engine of MSP as MSP.

\subsubsection{Contribution Prediction $I$}
The contribution prediction $I_{mn}$ from  MU $m$ to MSP $n$ is determined by the accuracy $\theta_{m}$ of the local model, the total amount of training data $\lfloor\frac{T}{\tau_{n}}\rfloor|X_{mn}|$, and the potential value $\omega_{mn}$ of local data. Prior work~\cite{Game} characterized model contributions using a logarithmic function of the amount of local training data. However, evaluating the contribution solely by the training data size is one-sided, as there may be a large amount of redundant data or the validity of the local model cannot be guaranteed. It is more practical to comprehensively evaluate the amount of training data, model accuracy, and the potential value of local data; thus, we denote $I_{mn}$ as
\begin{equation}
I_{mn}=\frac{\omega_{mn} \epsilon \ln(1+\eta \lfloor\frac{T}{\tau_{n}}\rfloor|X_{mn}|)}{\theta_{m}},
\end{equation}where $\lfloor\frac{T}{\tau_{n}}\rfloor$ is the number of iterations that the task of MSP $n$ can be performed within $T$. $\epsilon$ and $\eta$ are system parameters from experiments as obtained in~\cite{ln}. $\theta_{m}\in\left (0, 1 \right )$ characterizes the accuracy of local training as decided by the MU. Here, $\theta\rightarrow 0$ means that a high-accuracy local model is available, while $\theta\rightarrow 1$ implies that the accuracy of the local model is low. The local data potential value $\omega_{mn}$ describes the difference between model predictions and true labels, defined as
\begin{equation}
\omega_{mn}=\frac{1}{|X_{mn}|}\sum_{i=1}^{|X_{mn}|}(\hat z_{i}-z_{i})^{2},
\end{equation}where $\hat z_{i}$ denotes the inferred label from the current global model and $z_{i}$ is its ground-truth label. If the value of $\omega_{mn}$ is large, it indicates that the current global model is not working well on local data or there are unseen samples. Therefore, updating the model by training on these data may lead to better performance of the global model.
If the value of $\omega_{mn}$ is small, the current data has little potential to improve the global model further since the knowledge of the data has already been learned or the data size is small. Note that to shorten the decision time, we use the potential value of the initial local data to be a proxy for the average potential value of all data within $T$. This is a reasonable approach when $T$ is not too long.
\begin{figure}[t]
	\centering
	\includegraphics[width=\linewidth]{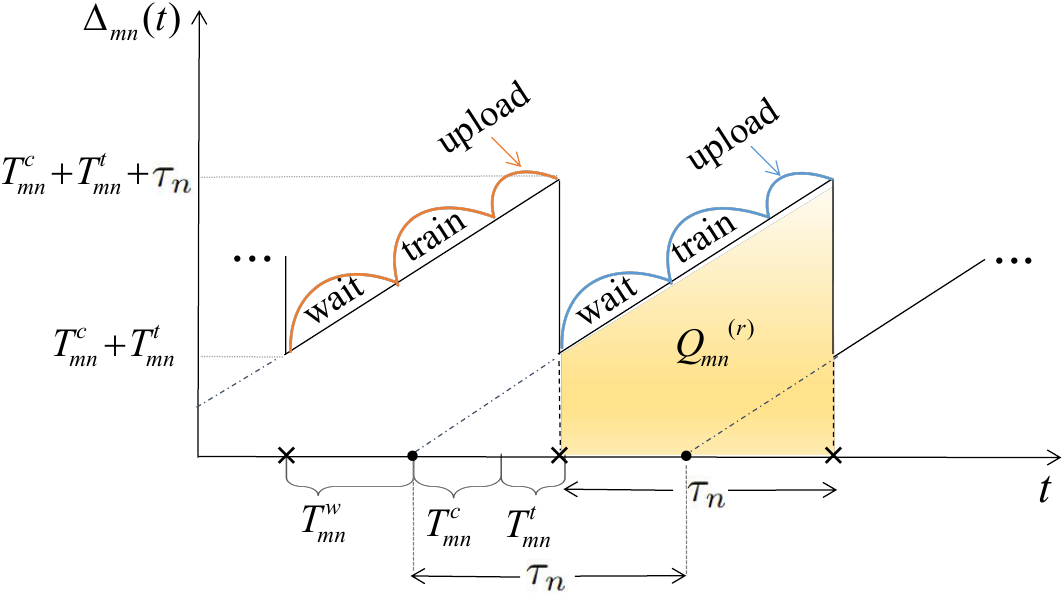}
	\caption{The AoI evolution of the local updates sent from MU $m$ to MSP $n$.}
	\label{aoi}
\vspace{-0.1in}
\end{figure}
\subsubsection{Age of Information $\Delta$}

The whole FL process involves model training, uploading, and waiting, which together constitute a complete cycle of model updating.
From the perspective of MSP $n$, the age evolution of the model contributed by MU $m$ is shown in Fig.~\ref{aoi}.
Here, the filled circles denote the time instances when MU $m$ starts training the local model. The intersections show the time instances when MSP $n$ receives the corresponding local updates, at which point the AoI drops to the lowest.  

We denote that at time $t$, the generation time of the last update received by MSP $n$ from MU $m$ is $s(t)$. Then, the instantaneous AoI of MU $m$'s local model measured at MSP $n$, $\Delta
_{mn}(t)$, is
\begin{equation}
    \Delta_{mn}(t)=t-s(t),
\end{equation}
which shows the elapsed time since the generation of the latest local updates. A smaller instantaneous AoI means that the latest local updates received are fresher at that time.

To obtain a comprehensive understanding of the overall freshness of model updates over a certain period, we use \textit{average} AoI to evaluate the freshness of the models.
With the aid of \figurename~\ref{aoi}, the time-averaged AoI of the local model from MU $m$ to MSP $n$ can be expressed as
\begin{equation}\label{ave_aoi}
\begin{aligned}
    \overline \Delta
_{mn}&=\mathop {\lim }\limits_{Z \to \infty } \frac{1}{Z}\int_0^Z {{\Delta _{mn}}(t)} \,dt=\mathop {\lim }\limits_{Z \to \infty }\sum_{r=1}^Z\frac{{Q_{mn}}^{(r)}}{Z}\\&= \frac{1}{2}\tau_{n}+T_{mn}^{c}+T_{mn}^{t},
\end{aligned}
\end{equation}
where ${Q_{mn}}^{(r)}$ is the $r$-th trapezium under the curve. A lower average AoI indicates that the local updates are generally fresher over a long period.

Since MU $m$ continuously collects data during the FL process, the amount of training data used in each round is $|X_{mn}|=x_{m}\tau_{n}$,
where $x_{m}$ denotes the data (number of floats) collected per unit time.
The cumulative time $T_{mn}^{c}$ for local training is determined by the amount of training data, computational resource $f_{mn}$ allocated to MSP $n$, and its local accuracy threshold $\theta_{m}$, which is equivalent to
\begin{equation}\label{Tc}
T_{mn}^{c}=\log(1/\theta_{m})\frac{ x_{m}\tau_{n}}{f_{mn}}.
\end{equation}
A smaller value of $\theta$ indicates higher accuracy, but leads to a higher MU cost, i.e., the number of local iterations, which is upper bounded by $\log (1/\theta_{m})$~\cite{lin2021friend}.

Furthermore, the upload time $T_{mn}^{t}$ depends on the size of local updates $b_{mn}$ (in bits) and the resource $B_{mn}$ (in Hz) allocated for communication between MU $m$ and MSP $n$ (i.e., uplink bandwidth), defined as
\begin{equation}\label{Tt}
T_{mn}^{t}= \frac{b_{mn}}{B_{mn}\log_{2}(1+\varsigma_{mn})},	
\end{equation}where $\varsigma_{mn}=\frac{p^{t}_{m}g_{mn}}{\sigma ^{2}}$ denotes the signal-to-noise ratio (SNR)~\cite{10924406} for the communication between MU $m$ and MSP $n$.  $g_{mn}$ and $p^{t}_{m}$ are the corresponding channel gain and the transmission power of MU $m$, respectively.  $\sigma^{2}$ is the power of additive white Gaussian noise (AWGN).

Finally, by substituting (\ref{Tc}) and (\ref{Tt}) into (\ref{ave_aoi}), $ \overline \Delta_{mn}$ can be expressed as
\begin{equation}
 \overline \Delta_{mn}= \frac{1}{2}\tau_{n}+\frac{ x_{m}\tau_{n}\log(1/\theta_{m})}{f_{mn}}+\frac{b_{mn}}{B_{mn}\log_{2}(1+\varsigma_{mn})}.
\end{equation}

\section{Game Formulation}\label{IV} 
We first define the utility functions of MU $m$ and MSP $n$ in Section~\ref{Game-MU} and Section~\ref{Game-MSP}, respectively. Then, we formulate the interactions among MUs and MSPs as an equilibrium problem with equilibrium constraints (EPEC) in Section~\ref{Game-EPEC}, where the equilibrium criterion exists at both the level of MUs and MSPs due to the conflicting interests between them.
\subsection{Utility of MU}\label{Game-MU}
We define the utility of MU $m$ composed of the rewards from MSPs and the cost incurred for contribution, which is also affected by resource constraints and immersion requirements for consuming vehicular metaverse services. The cost incurred due to local training and model uploading can be expressed as
\begin{equation}
C_{mn}=c_{m}^{f}\log(1/\theta_{m})f_{mn}+c_{m}^{B}B_{mn},
\end{equation}where $c_{m}^{f}$ and $c_{m}^{B}$ denote the cost factors of computational and communication resources, respectively. Let $\bm{f}_{m}\triangleq\left( f_{m1}, f_{m2}, \ldots, f_{mN}\right)$ and $\bm{B}_{m}\triangleq\left( B_{m1}, B_{m2}, \ldots, B_{mN}\right)$ be the computational and communication allocation profiles of MU $m$, respectively. Then, the utility maximization problem for MU $m$ within $T$ can be formulated as
\newtheorem{problem}{Problem}
\begin{problem}
    \begin{equation}\label{pro1c}
\begin{aligned}
&\max\, \, \Phi_{m}(\bm{f}_{m}, \bm{B}_{m})=\sum_{n=1}^N(p_{mn}V_{mn}-C_{mn})\\
\mathrm{s.t.}\  &\ C1:\sum_{n=1}^Nf_{mn} <  f_{m}^{max}, \,\, \sum_{n=1}^NB_{mn} <  B_{m}^{max},\\
&\ C2:\log(1/\theta _{m})\frac{ x_{m}\tau_{n}}{f_{mn}}+\frac{b_{mn}}{B_{mn}\log_{2}(1+\varsigma_{mn})}\leq \frac{1}{2}\tau_{n},\\
&\ C3:\frac{S_{m}}{f^{max}_{m}-\sum_{n=1}^Nf_{mn}}<  T^{req},
\end{aligned}
\end{equation}
\end{problem}
\noindent where $p_{mn}$ is the reward offered by MSP $n$ to MU $m$ for their contribution. $S_{m}$ indicates the minimum computational resource required to enjoy other basic services.  
Constraints C1 and C3 indicate that the MU's total computational and communication resources are limited and cannot be fully used to contribute local models, given the MU's demand for basic services.
Constraint C2 ensures that $V_{mn}$ is non-negative, guaranteeing that the value derived from the task meets the MSP's minimum requirements. Specifically, the total time for local training and uploading in each round cannot exceed the time constraint $\frac{1}{2}\tau_{n}$.
The goal of each MU is to choose the optimal allocation of computational $\bm{f}_{m}$ and communication $\bm{B}_{m}$ resources to maximize its utility. 

\subsection{Utility of MSP}\label{Game-MSP}
Each MSP has a gain function $\psi$ associated with the IoM to quantify the cumulative benefit of all local models contributed by MUs. In this paper, we adopt $\psi_{n} =\mu_{n} \ln(1+\sum_{m=1}^MV_{mn})$ as a monotonically increasing, differentiable, strictly concave IoM function, which is a simplification of the widely adopted function~\cite{8758205}.
Let $\bm{p}_{n}\triangleq\left ( p_{1n}, \ldots, p_{Mn} \right )$ and $\bm{p}_{\textrm{-}n}$ represent the reward profile of MSP $n$ and all other MSPs except MSP $n$, respectively. Then, the optimization problem for the MSP is defined as
\begin{problem}
    \begin{equation}\label{pro2}
\begin{aligned}
&\max \, \, \Psi_{n}(\bm{p}_{n},\bm{p}_{\textrm{-}n})=\psi_{n}-\sum_{m=1}^Mp_{mn}V_{mn} ,\\
&\mathrm{s.t.}\   p_{mn}>0, n\in \mathcal{N}, m \in \mathcal{M},\\
\end{aligned}
\end{equation}
\end{problem}
\noindent where $\mu_{n}$ refers to the profit conversion coefficient~\cite{profit} from IoM, which is adjusted according to the AR services offered by the different MSPs.
Note that the other MSPs' decisions $\bm{p}_{\textrm{-}n}$ are captured by the MUs' responses. 
Moreover, when MSP $n$ determines its reward
$\bm{p}_{n}$ for different MUs, the MSP needs to consider the
rewards offered by other MSPs (i.e., $\bm{p}_{\textrm{-}n}$ ) as well as
the strategies of all MUs (i.e., $(\bm{f}_{m},\bm{B}_{m}), m \in \mathcal{M}$). This thereby leads to 
reward competition among MSPs. Therefore, the optimization among MSPs can be considered a non-cooperative game, termed a multi-MSP rewarding game $\Omega$, as follows.
\begin{myDef}
    A multi-MSP rewarding game $\Omega $ is a tuple $\Omega =\left \{ \mathcal{N}, \bm{p}, \bm{\Psi}\right \}$ defined by 
    \begin{itemize}{
		\item {Players: The set of MSPs; }
		
		\item { Strategies: The reward decisions $\bm{p}_{n}$ of any MSP $n$;}
		
		\item { Utilities: The vector $\bm{\Psi}=\left \{  \Psi_{1},  \Psi_{2}, \ldots,  \Psi_{n}\right\}$
			contains the utility functions of all the MSPs defined in (\ref{pro2})}.
	}
\end{itemize}
\end{myDef}
\begin{figure}[t]
	\centering
	\includegraphics[width=\linewidth]{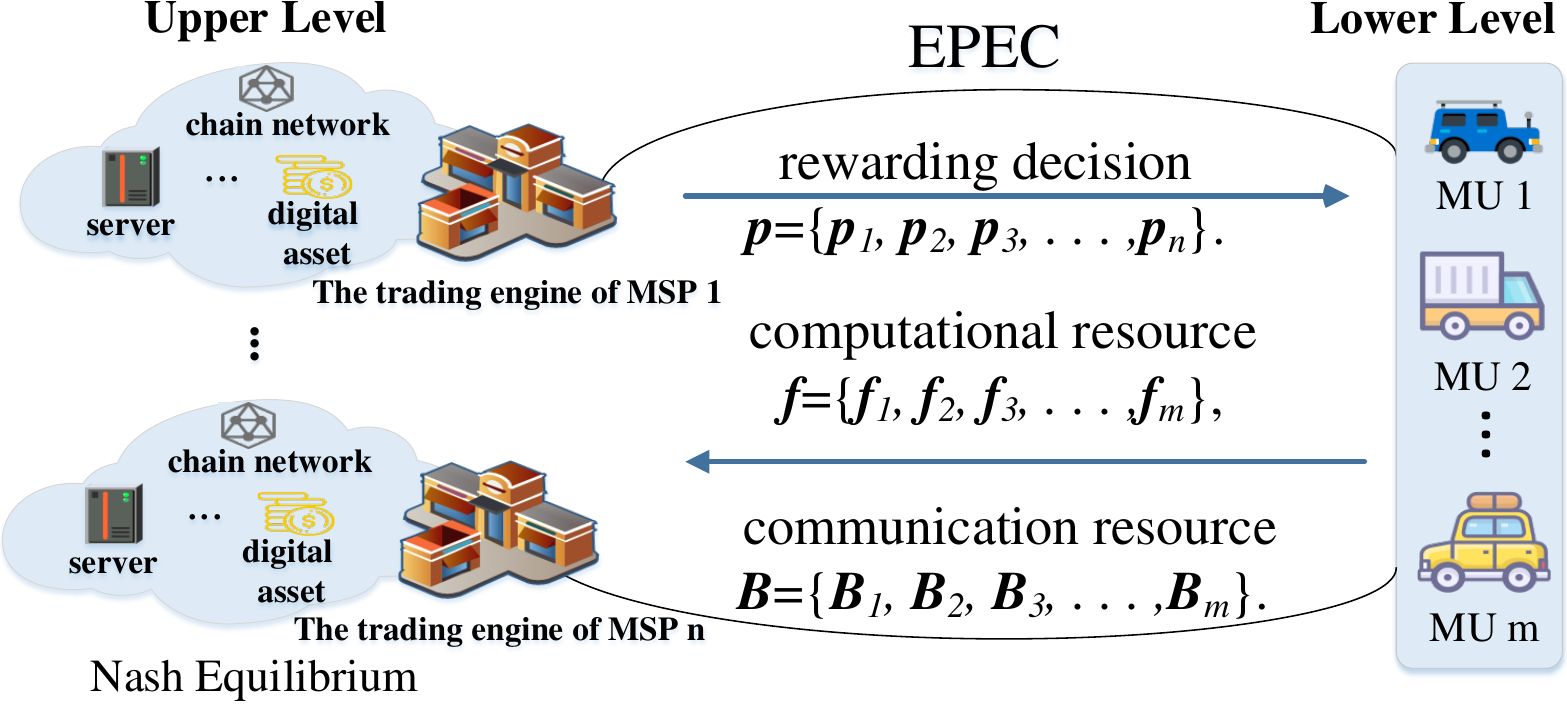}
	\caption{The hierarchical structure of equilibrium problem with equilibrium constraints (EPEC).}
	\label{game}
\vspace{-0.1in}
\end{figure}
\subsection{Multi-MSP Multi-MU Game as EPEC}\label{Game-EPEC}

As shown in Fig.~\ref{game}, MUs and MSPs negotiate allocation strategies of computational and communication resources and
reward decisions to maximize their benefits.
In the upper level, MSPs determine the rewards they are willing to offer by considering their cost, the responses of MUs, and the decisions of other MSPs. In the lower level, each MU $m$ gives the optimal allocation response for computational and communication resources by considering its
resource constraints, cost, and rewards from different MSPs. 

The objective of EPEC is to find the equilibria at two levels, i.e., the point at which the MSPs' (leaders') utilities are maximized given that the MUs (followers) will choose their best responses. For the proposed EPEC, the equilibria at two levels are defined as follows.
\begin{myDef}
    Let $(f^{*}_{mn}, B^{*}_{mn})$ and $p^{*}_{mn}$ denote the optimal computational and communication resource allocation of MU $m \in \mathcal{M}$ and the optimal reward decision of MSP $n \in \mathcal{N}$, respectively. Then, the points $(f^{*}_{mn}, B^{*}_{mn})$ and $p^{*}_{mn}$ are the equilibria at two levels if the following conditions hold:
\begin{equation}
\begin{aligned}
&\Phi_{m}((\bm{f}_{m}^{*},\bm{B}_{m}^{*}),\bm{p}_{n}^{*})\geq \Phi_{m}((\bm{f}_{m},\bm{B}_{m}),\bm{p}_{n}^{*}), \forall m \in \mathcal{M},\\
&\Psi_{n}\left ( \bm{f}_{m}^{*}(\bm{p}_{n}^{*},\bm{p}_{-n}^{*}), \bm{B}_{m}^{*}(\bm{p}_{n}^{*},\bm{p}_{-n}^{*}) ,\bm{p}_{n}^{*},\bm{p}_{-n}^{*}\right ) \\&\geq\Psi_{n}\left ( \bm{f}_{m}^{*}(\bm{p}_{n}^{*},\bm{p}_{-n}^{*}), \bm{B}_{m}^{*}(\bm{p}_{n}^{*},\bm{p}_{-n}^{*}), \bm{p}_{n},\bm{p}_{-n}^{*}\right ), \forall n\in \mathcal{N},
\end{aligned}
\end{equation}
where $\bm{p}_{-n}^{*}$ denotes the optimal reward vector for all MSPs except $n$.
\end{myDef}
In summary, the MSPs' optimization problems are
formulated as the following EPEC problems:
 \begin{equation}\label{pro-epec}
\begin{aligned}
& \max_{p_{n}} \, \, \Psi_{n}=\psi_{n}-\sum_{m=1}^M \bigg[ p_{mn}I_{mn}\bigg( \frac{1}{2}\tau_{n}-\frac{ x_{m}\tau_{n}\log(1/\theta_{m})}{f_{mn}^{*}} \\
& \qquad \qquad \qquad \qquad \qquad -\frac{b_{mn}}{B_{mn}^{*}\log_{2}(1+\varsigma_{mn})}\bigg)\bigg], \\
& \mathrm{s.t.} \  \left\{ \begin{aligned}
& p_{mn}>0,\   n\in \mathcal{N},\  m \in \mathcal{M},\\
& (\bm{f}_{m}^{*},\bm{B}_{m}^{*}) = \arg\max \Phi_{m}(\bm{f}_{m}, \bm{B}_{m}), \\
& \qquad\qquad\quad\quad\mathrm{s.t.} \  C1, C2, C3.
\end{aligned} \right.
\end{aligned}
\end{equation}
To investigate the above EPEC, we address the lower level (\textbf{Problem 1}) and the upper level (\textbf{Problem 2}) by using the backward induction methods in the following section.

\section{EPEC Analysis and Solutions}\label{V}
In this section, we utilize backward induction to analyze the EPEC formulated in~\eqref{pro-epec}. Specifically, we prove the existence and uniqueness of the equilibrium solutions at two hierarchical levels: (1) the optimal allocation of computational and communication resources for MUs (lower level), and (2) the optimal reward decisions for MSPs (upper level).

\subsection{Lower Level: Optimal Resource Allocation for MUs}
In the lower level of EPEC, for any reward decisions $\bm{p}$ given by MSPs, MU $m$ aims to solve~\textbf{Problem~1} in~\eqref{pro1c} to determine its optimal computational and communication resource allocations, i.e., $\bm{f}_{m}^{*}$ and $\bm{B}_{m}^{*}$, to maximize its utility. Below, we analyze and derive the unique optimal allocation.

\begin{myTheo}\label{the1}
     \textbf{Problem 1} is 
strictly concave and has a unique globally optimal solution in the lower level. That is, for each MU $m$, there exists a unique resource allocation tuple $(\bm{f}_{m}^{*}, \bm{B}_{m}^{*})$ that maximizes its utility.
\end{myTheo}
    \begin{proof}
	We first examine the Hessian matrix of MU $m$'s utility $\Phi_{m}(\bm{f}_{m},\bm{B}_{m})$ with respect to $(f_{mn},B_{mn})$. Let this Hessian matrix be denoted by 
$\bm{H}_{m}$, which can be block-diagonalized as
		\begin{equation}
	\bm{H}_{m}=\begin{bmatrix}
	\bm{H}_{m}^{f} &0 \\ 
	0&\bm{H}_{m}^{B} 
	\end{bmatrix}.
	\end{equation}
    
    The block matrix $\bm{H}^{f}_{m}$ can be computed as the second-order partial derivative of $\Phi_{m}(\bm{f}_{m},\bm{B}_{m})$ with respect to $f_{mn}$, i.e.,
	\begin{equation}
	\begin{aligned}
	\bm{H}^{f}_{m}&=\left [  \frac{\partial^{2}\Phi_{m}(\bm{f}_{m}, \bm{B}_{m})}{\partial f_{mn}\partial f_{mn'}}\right ]_{n,n'\in \mathcal{N}}\\&= -diag\left(h^{f}_{m1}, h^{f}_{m2} \ldots,h^{f}_{mN}\right)<\bm{0},
	\end{aligned}
	\end{equation}where $h^{f}_{mn}= p_{mn}I_{mn}\frac{2 x_{m}\tau_{n}}{f_{mn}^{3}}\log(1/\theta_{m})$.
	Similarly, the block matrix $\bm{H}^{B}_{m}$ can be calculated by 
	\begin{equation}
	\begin{aligned}
	\bm{H}^{B}_{m}&=\left [\frac{\partial^{2}\Phi_{m}(\bm{f}_{m}, \bm{B}_{m})}{\partial B_{mn}\partial B_{mn'}}\right]_{n,n'\in \mathcal{N}}\\&= -diag\left(h^{B}_{m1}, h^{B}_{m2}, \ldots, h^{B}_{mN}\right)<\bm{0},
	\end{aligned}
	\end{equation}where $h^{B}_{mn}= \frac{2b_{mn}p_{mn}I_{mn}}{B_{mn}^{3}\log_{2}(1+\varsigma_{mn})}$.
 
    Since $\bm{H}^{f}_{m}$ and $\bm{H}^{B}_{m}$ are diagonal matrices with strictly negative diagonal elements, they are negative definite. Consequently, the block-diagonal Hessian $\bm{H}_{m}$ is negative definite, implying $\Phi_{m}(\bm{f}_{m},\bm{B}_{m})$ is strictly concave and continuous. Moreover, the objective function $\Phi_m$ tends to $-\infty$ as any variable approaches the boundary of the feasible set (e.g., $f_{mn} \to 0^+$ and $B_{mn} \to 0^+$, or $f_{mn} \to \infty$ and $B_{mn} \to \infty$). Hence, the maximum is necessarily attained in the interior. Combined with strict concavity over the convex feasible region defined by constraints C1-C3, this ensures that \textbf{Problem 1} admits a unique global optimum $(\bm{f}_m^*, \bm{B}_m^*)$ (see Section IV in~\cite{boyd2004convex}).
\end{proof}

Consequently, MU $m$ has a unique best-response resource allocation for each MSP by solving \textbf{Problem 1}. From the Karush–Kuhn–Tucker (KKT) conditions, we derive the following proposition.

\textbf{Proposition 1}. Given the digital currency price offered by the MSPs and the Lagrange multipliers induced by the constraints, the optimal computational resource $f_{mn}$ allocated for MSP $n$ by MU $m$ satisfies:
\begin{equation}
f_{mn}^{*}=\begin{cases}
\sqrt{\frac{p_{mn} I_{mn} x_{m}\tau_{n}}{ c_{m}^f}}, &p_{mn} \ge \frac{F_{mn}^{2}}{I_{mn}},\\ 
F_{mn}\sqrt{\frac{x_{m}\tau_{n}}{ c_{m}^f}},  &0<p_{mn}<\frac{F_{mn}^{2}}{I_{mn}},\\ 0, &\text{otherwise},
\end{cases}
\end{equation} and the optimal communication bandwidth $B_{mn}$ is:
\begin{equation}
B_{mn}^{*}=\begin{cases}
\sqrt{\frac{p_{mn} I_{mn}b_{mn}}{ c_{m}^B \log_{2}(1+\varsigma_{mn})}}, &p_{mn} \ge \frac{F_{mn}^{2}}{I_{mn}}, \\ 
F_{mn}\sqrt{\frac{b_{mn}}{ c_{m}^B \log_{2}(1+\varsigma_{mn})}},  &0<p_{mn}<\frac{F_{mn}^{2}}{I_{mn}},\\ 0, &\text{otherwise},
\end{cases}
\end{equation}where $F_{mn}=2\log(1/\theta_{m}) \sqrt{\frac{x_{m}c_{m}^{f}}{\tau_{n}} } +\frac{2\sqrt{b_{mn}c_{m}^{B}}}{\tau_{n}\sqrt{\log_{2}(1+\varsigma_{mn})} }$. For the detailed proof of \textbf{Proposition 1}, please refer to \textbf{Appendix A}.

According to \textbf{Proposition 1}, the optimal strategy for each MU $m$ is influenced by five main factors: $p_{mn}$, $I_{mn}$, $c_{m}^{f}$, $c_{m}^{B}$ and $\tau_{n}$. Specifically, MU $m$ focuses on the rewards paid by MSPs, and the MU tends to put more computational and communication resources into the contribution when $p$ increases.  Moreover, MUs are more willing to invest more resources when the potential contribution prediction $I_{mn}$ is higher, resulting in greater benefits. In addition, the virtual deadline $\tau_{n}$ set by the MSP $n$ determines the minimum criteria for resource allocation.

\subsection{Upper Level: Optimal Reward Equilibrium among MSPs}
In the upper level of the EPEC, each MSP $n$ competes with other MSPs and determines its reward vector $\bm{p}_{n}$. Specifically, given the responses $(\bm{f},\bm{B})$ from all MUs, and other MSPs' decisions $\bm{p}_{\textrm{-}n}$, the optimal reward decision $\bm{p}_{n}$ of MSP $n$ can be obtained by solving \textbf{Problem 2}, defined as follows.

\textbf{Proposition 2}.
    Given other MSPs' reward vectors $\bm{p}_{\textrm{-}n}$,
the optimal strategy of MSP $n$ is
\begin{equation}
\begin{aligned}
\bm{p}_{n}^{*}=arg\max_{p_{mn}>0}\Psi_{n}\left ( \bm{f}_{n}^{*}(\bm{p}_{n},\bm{p}_{-n}), \bm{B}_{n}^{*}(\bm{p}_{n},\bm{p}_{-n}) ,\bm{p}_{-n}\right ).
\end{aligned}
\end{equation}

Then, we analyze the existence and uniqueness of the optimal reward decision equilibrium in \textbf{Theorem~\ref{concave2}}.

\begin{myTheo}\label{concave2}
    There exists a unique Nash equilibrium in the multi-MSP rewarding game $\Omega$, ensuring a unique optimal reward decision profile $\{\bm{p}_{n}^{*}\}_{n\in \mathcal{N}}$.
\end{myTheo}
\begin{proof}
	We define the Hessian matrix of $\Psi_{n}$ with respect to its reward vector $\bm{p}_{n}$ as $(\bm{\Lambda}_{n}+\bm{H}_{n})$. The matrix $\bm{\Lambda}_{n}=diag \left(\frac{\partial^{2} \Psi_{n}}{\partial p_{1n}^{2}},\ldots,\frac{\partial^{2} \Psi_{n}}{\partial p_{Mn}^{2}} \right)$ and
	the second-order partial derivative matrix $\bm{H}_{n}$ is expressed by
	\begin{equation}
	\bm{H}_{n}=\begin{bmatrix}
	0 &  \frac{\partial^{2} \Psi_{n}}{\partial p_{1n}\partial p_{2n}}& \cdots  & \frac{\partial^{2} \Psi_{n}}{\partial p_{1n}\partial p_{Mn}}\\ 
	\frac{\partial^{2} \Psi_{n}}{\partial p_{2n}\partial p_{1n}} & 0 & \cdots  & \frac{\partial^{2} \Psi_{n}}{\partial p_{2n}\partial p_{Mn}}\\ 
	\vdots & \vdots  &  \ddots & \vdots \\ 
	\frac{\partial^{2} \Psi_{n}}{\partial p_{Mn}\partial p_{1n}}&  \frac{\partial^{2} \Psi_{n}}{\partial p_{Mn}\partial p_{2n}}&\cdots   & 0
	\end{bmatrix},
	\end{equation}where 
	\begin{equation}\label{last-VV}
\begin{aligned}
\frac{\partial^{2} \Psi_{n}}{\partial p_{mn}^{2}}=&\ \mu_{n}\frac{V_{mn}''(1+\sum_{m=1}^MV_{mn})-(V_{mn}')^{2}}{(1+\sum_{m=1}^MV_{mn})^{2}}\\&-2V_{mn}'-p_{mn}V_{mn}''<0, \   \forall m \in \mathcal{M},
\end{aligned}
\end{equation}
\begin{equation}
\begin{aligned}
&\frac{\partial^{2} \Psi_{n}}{\partial p_{mn}\partial p_{m'n}}=-\mu_{n}\frac{V_{mn}'V_{m'n}'}{(1+\sum_{m=1}^MV_{mn})^{2}}<0,  \\ &\forall m \in \mathcal{M},\ m\neq m'.
\end{aligned}
\end{equation}

 We randomly choose a vector $\bm{h}\in \mathbb{R}^{M\times 1}$ with elements not all $0$. Then, we have $\bm{h}^{T}(\bm{\Lambda}_{n}+\bm{H_{n}})\bm{h}=\sum_{m=1}^M (h^m)^2 (\frac{\mu_n V_{mn}''}{1+\sum_{m=1}^MV_{mn}} - 2V_{mn}' - p_{mn}V_{mn}'' )-\frac{\mu_n}{(1+\sum_{m=1}^MV_{mn})^2} ( \sum_{m=1}^M h^m V_{mn}' )^2 <0$, indicating that the utility function $\Psi_{n}$ is strictly concave. According to ~\cite{10415185}, there exists a unique Nash equilibrium in the multi-MSP rewarding game $\Omega$. 
The proof of $\frac{\partial^{2} \Psi_{n}}{\partial p_{mn}^{2}}<0$, $\frac{\partial^{2} \Psi_{n}}{\partial p_{mn}\partial p_{m'n}}<0$, and $\bm{h}^{T}(\bm{\Lambda}_{n}+\bm{H_{n}})\bm{h}<0$ can be found in \textbf{Appendix B}.
\end{proof}
 \begin{algorithm}[t]
\label{alo1}
	\caption{Entire Process of the Framework.}
	\LinesNumbered 
	\textbf{Input}: Action threshold $[p_{min}, p_{max}]$ and number of episodes $\mathcal{T}$\;
	\textbf{Initialization}: local observation $s_{n}$, actor network $\bm{\alpha}_{n}$, critic network $\bm{\beta}_{n}$, and episode buffer $\mathcal{D}_{n}$ for each MSP agent $n \in \mathcal{N}$\; 
    \BlankLine
    \textbf{\# Trading process using MDDR}: \\
	\For{episode = $1,2,\ldots, \mathcal{T}$}
	{\textbf{Concurrently for each MSP agent $n \in \mathcal{N}$:}
    \\ \For{epoch $k = 1,\ldots,|\mathcal{D}_{n}|$}
		{Observe space $s_{n}(k)$\; Choose price $\bm{p}_{n}(k) \in [p_{min}, p_{max}]$ by sampling from its current policy $\pi_{\alpha_{n}}(s_{n}(k))$\; Broadcast reward decisions to MUs\; Interact with the environment and receive responses from the MUs\; Calculate utility according to (\ref{pro2}) \; Store transition $e_{n}(k)=[s_{n}(k), \bm{p}_{n}(k), \mathcal{U}_{n}(k), s_{n}(k+1)]$\;}
		Update actor network $\bm{\alpha}_{n}$ and critic network $\bm{\beta}_{n}$ for each MSP agent $n$\;
		Clear the episode buffer $\mathcal{D}_{n}$ for each MSP agent $n$\;}
        \textbf{Return}: Reward decisions $\bm{p}$ of MSPs and resource allocation $[\bm{f}, \bm{B}]$ of MUs\;
    \BlankLine
   \textbf{\# FL training guided by trading results within time $T$}:\\
  { \ForEach{MSP agent $n \in \mathcal{N}$}{\ForEach{MU $m \in \mathcal{M}$}{
  			Based on the trading results, utilize $f_{mn}$ and $B_{mn}$ to perform the local FL task for MSP $n$, and then upload the local updates to MSP $n$. }  Aggregate received local updates and issue an updated global model\;
Supply the updated model to the AR engine\;}
 }   
\end{algorithm} 

Utilizing \textbf{Theorem~\ref{the1}} and \textbf{Theorem~\ref{concave2}}, we prove the existence and uniqueness of the hierarchical equilibrium in the proposed EPEC model. Concretely, each MU has a unique best-response resource allocation strategy in reaction to any given MSP rewards. At the same time, each MSP determines unique optimal pricing to maximize its own utility, given the rewards of other MSPs.

Although the backward induction approach theoretically ensures an equilibrium solution to \eqref{pro-epec}, solving it in real-world vehicular metaverse environments can be challenging due to dynamic network conditions and limited system knowledge (e.g., private cost parameters). In the following section, we address these practical issues by introducing a fully distributed reinforcement learning approach, enabling MSPs to adaptively and privately optimize their reward decisions in real time.

\section{Dynamic Reward Strategy for MSPs}\label{VI}
According to the EPEC proven above, it is possible to obtain an optimal solution. However, the optimal approach faces the following practical challenges: (i) In a time-varying network, the optimal approach is time-consuming due to the complexity of (\ref{pro2}), which is a non-linear problem with a complicated structure. Additionally, the reward decisions of MSPs are tightly coupled.
(ii) MSPs must have prior knowledge of  private information about MUs (e.g., cost factors) to determine their rewards, which raises privacy concerns for MUs. 
To address these challenges, we further formulate the multi-MSP rewarding game as a multi-agent Markov decision process (MAMDP)~\cite{8758205}, adapting to dynamic channel conditions. In this framework, each MSP is modeled as an individual agent that makes intelligent reward decisions.

\subsection{MAMDP for Multi-MSP Rewarding Game $\Omega$}  
 We train the reward model based on the state-of-the-art policy gradient method proximal policy optimization (PPO) for the reasons described in \cite{PPO_reason}.
The $MAMDP=\left \langle \mathcal{S}_{n}, \mathcal{A}_{n}, \mathcal{P}_{n}, \mathcal{U}_{n}\right \rangle$ for multi-MSP rewarding game is composed of state space $\mathcal{S}_{n} \triangleq \left \{s_{n} \right \}$, action space $\mathcal{A}_{n} \triangleq \left \{\bm{p}_{n} \right \}$, state transition probability $\mathcal{P}_{n} \triangleq \left \{P_{n} \right \}$, and utility $\mathcal{U}_{n} \triangleq \left \{ \Psi_{n} \right \}$.
For the MSP agent, the local observation contains the channel information and the responses of MUs, all of which are captured in the IoM of (\ref{iom}). 
Thus, we set the local observation of MSP $n$ at the $k$-th stage game defined as $s_{n}(k)=[V_{1n}(k),V_{2n}(k), \ldots, V_{mn}(k)]$, and the state of the environment is $\mathcal{S}(k)=[s_{1}(k), \ldots, s_{n}(k)]$.  
  At the $k$-th stage game, MSP $n$ observes a state $s_{n}(k)$ and determines an action $\bm{p}_{n}(k)$ within $[p_{min}, p_{max}]$. When an action $\bm{p}_{n}(k)$ is applied to state $s_{n}(k)$, the agent $n$ receives a utility $\mathcal{U}_{n}(k)$ from the environment. Taking into account the competition among agents, we define the reward $\mathcal{U}_{n}$ for each MSP by (\ref{pro2}).
 The state transition probability $P(s_{n}(k+1)\mid s_{n}(k),\bm{p}_{n}(k))$ leads to the new state $s_{n}(k+1)$ after executing an action $\bm{p}_{n}(k)$ at the
state $s_{n}(k)$.

\begin{figure}[!t]
	\centering
	\includegraphics[width=\linewidth]{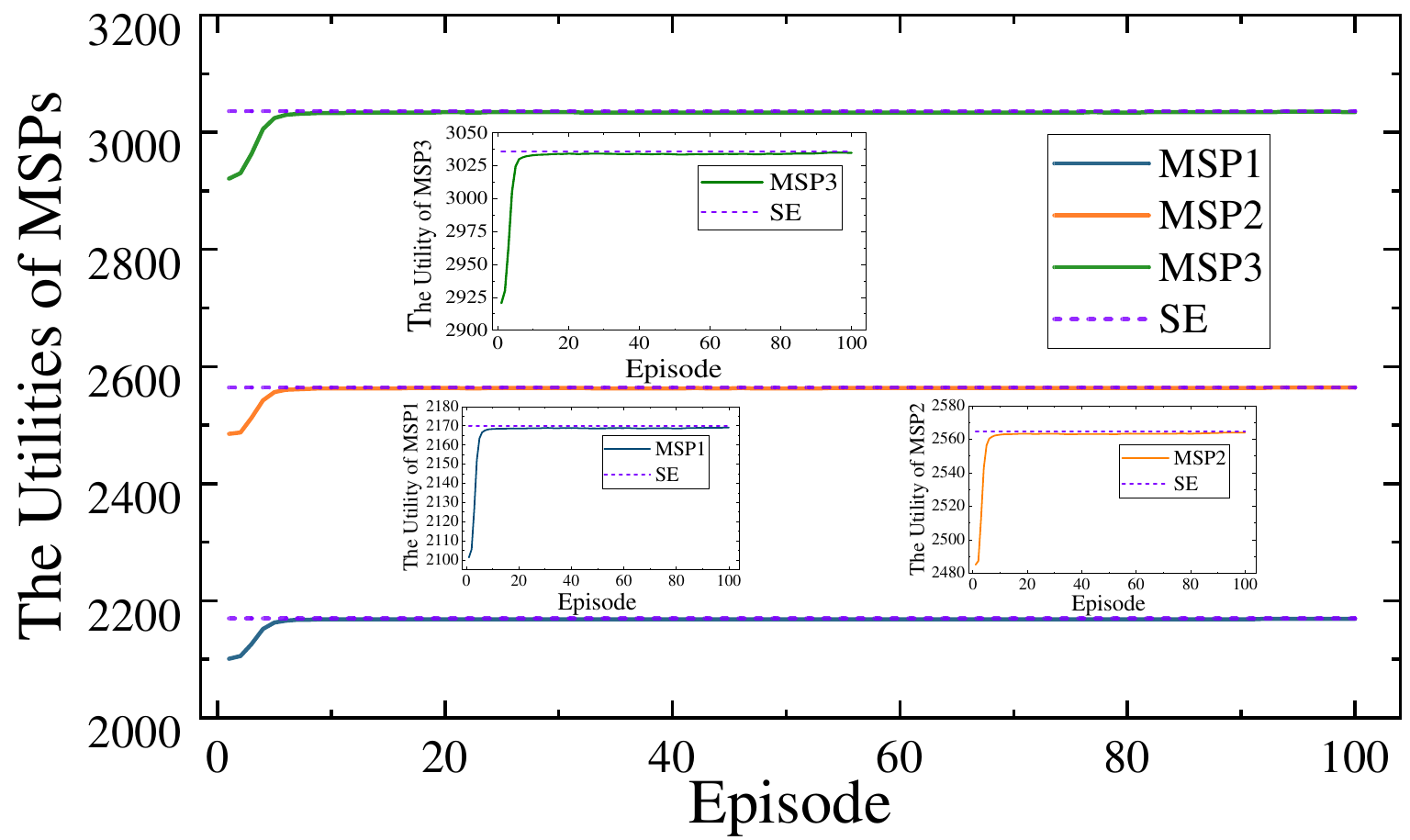}
	\caption{The utilities of MSPs.}
	\label{optimal}
 \vspace{-0.1in}
\end{figure}
\begin{figure*}[!t]
	\centering
	\subfigure[The utilities of MUs] {
		\label{MU}
		\includegraphics[scale=0.14]{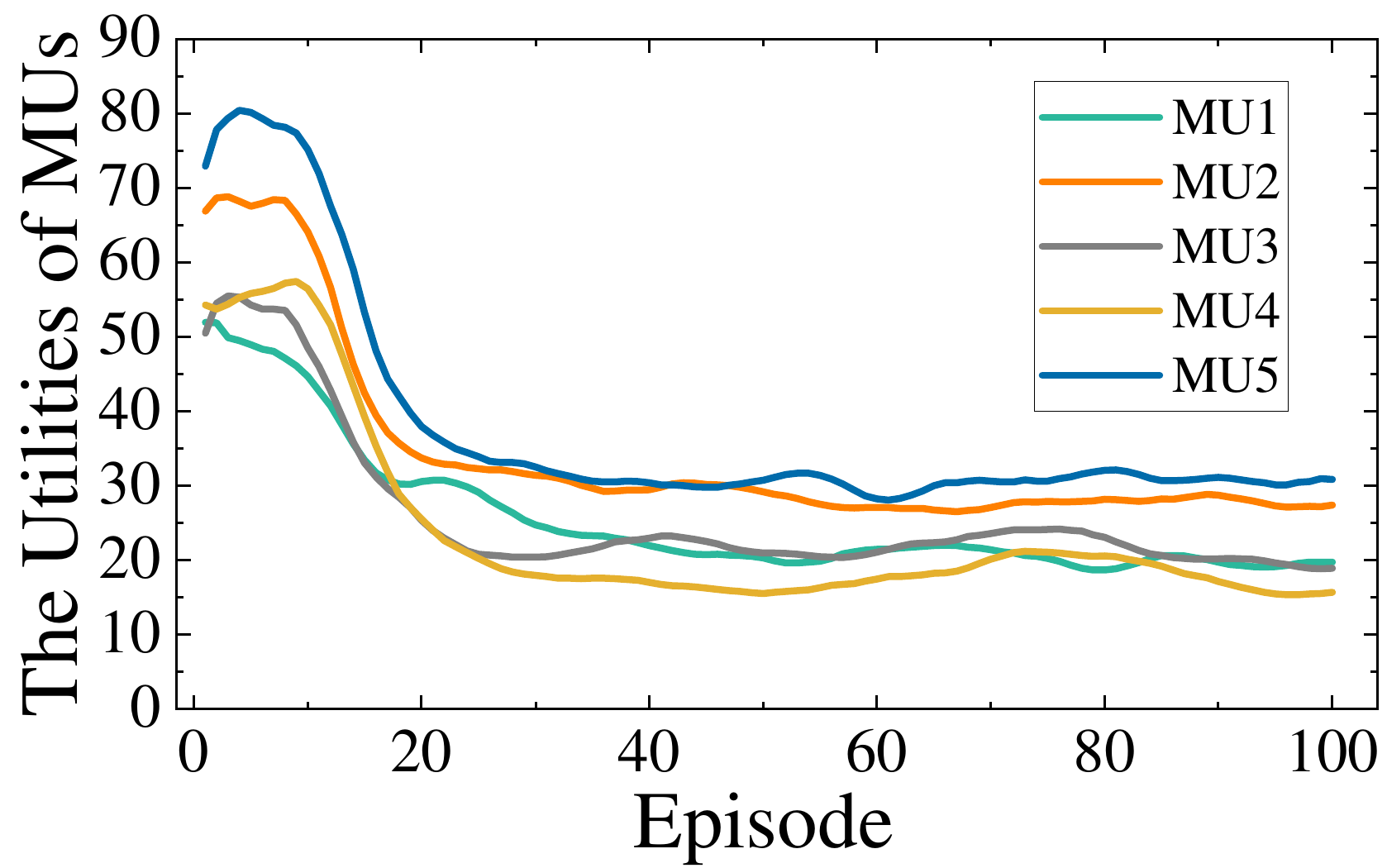}
	}
	\subfigure[The reward decisions of MSP 1] {
		\label{MSP1_price}
		\includegraphics[scale=0.14]{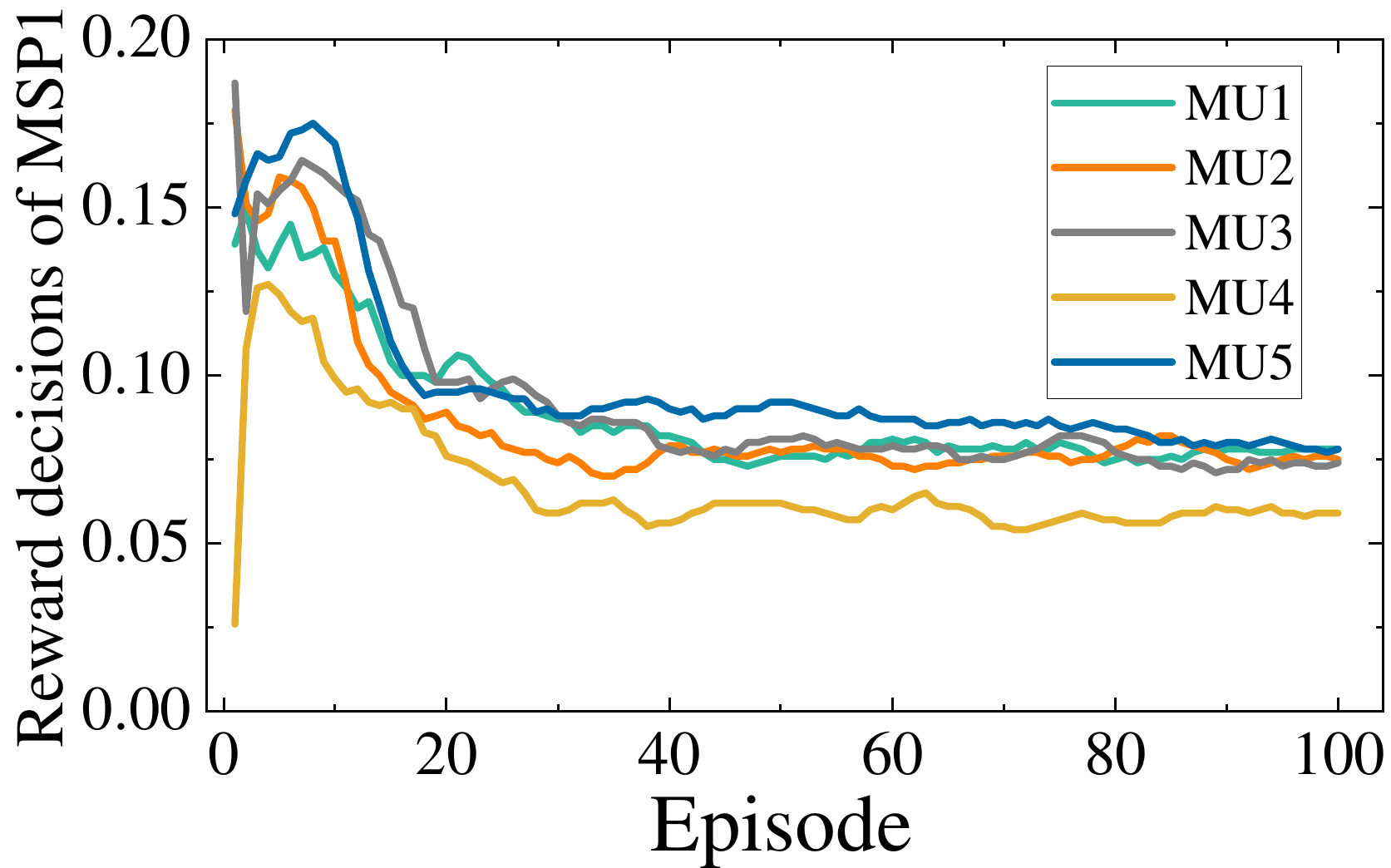}
	}
	\subfigure[The reward decisions of MSP 2] {
		\label{MSP2_price}
		\includegraphics[scale=0.14]{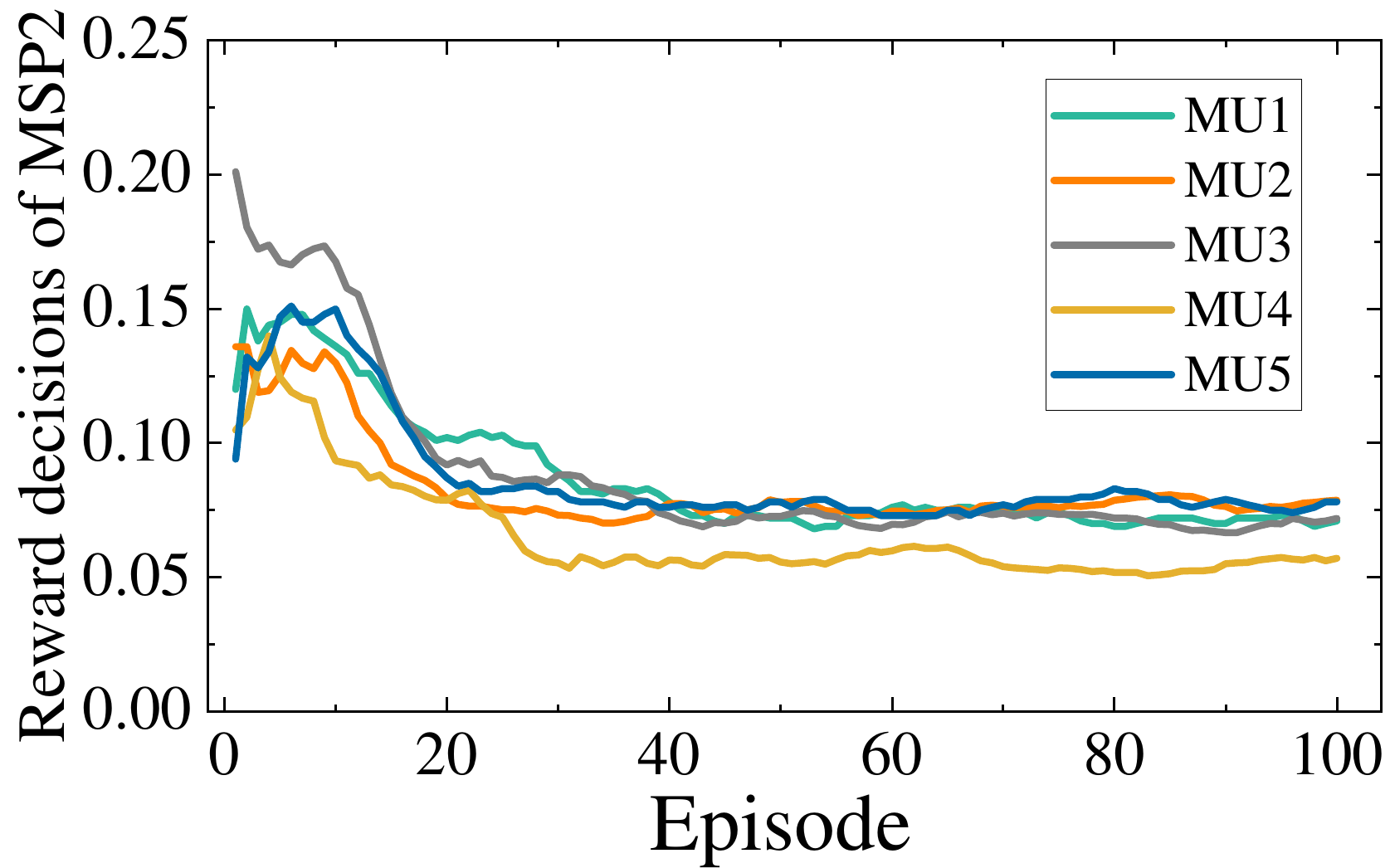}
	}
	\subfigure[The reward decisions of MSP 3] {
		\label{MSP3_price}
		\includegraphics[scale=0.14]{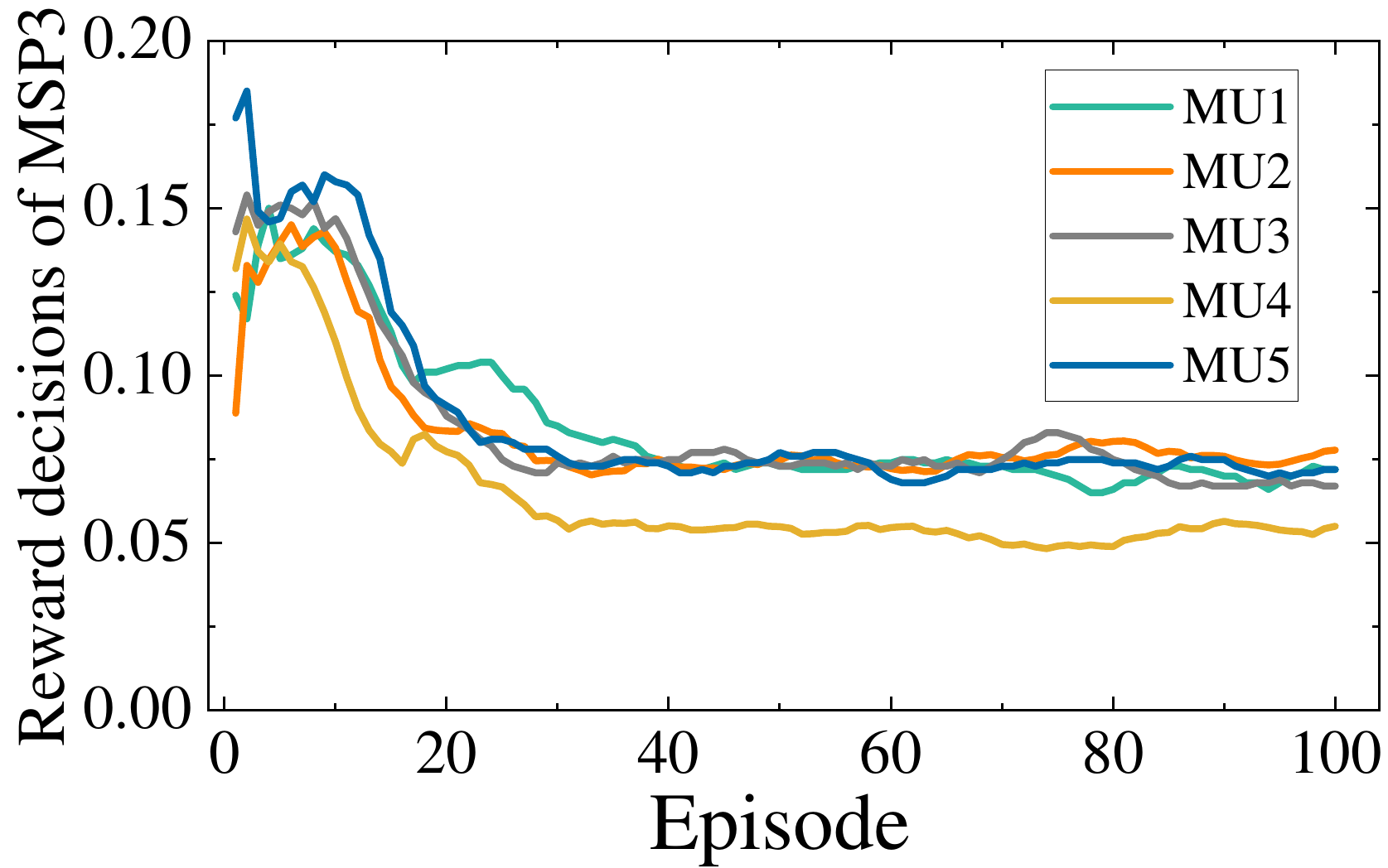}
	}
	\caption{The convergence processes of MDDR under dynamic networks.}
	\label{DRL}
 \vspace{-0.1in}
\end{figure*}
\subsection{The Multi-agent DRL-based Dynamic Reward (MDDR)}
In our work, each MSP agent operates the DRL-based dynamic reward algorithm in a fully distributed manner. 
\textbf{Algorithm \ref{alo1}} shows the pseudocode for model trading using MDDR under a dynamic environment (Lines 4-17) and the guidance for the FL process (Lines 19-25).

\textbf{Trading process:} At the start of the game, each MSP $n$ initializes its actor $\bm{\alpha}_{n}$ and critic network $\bm{\beta}_{n}$, and episode buffer $\mathcal{D}_{n}$ for each MSP agent $n$ (Line 2).
Each agent $n$ feeds the observed information into the policy network $\bm{\alpha}_n$ to derive its reward policy (Lines 7-8).
Subsequently, each MSP agent $n$ broadcasts its reward decisions, interacts with the environment, receives feedback from the MUs, calculates its utility, and stores the experience in the episode buffer $\mathcal{D}_n$ (Lines 9-12).
The experience is collected through this interactive process (Lines 7-12) until the buffer $\mathcal{D}_{n}$ is full. Once the buffer is full, the critic network $\bm{\beta}_n$ evaluates the actor network $\bm{\alpha}_n$ based on the collected experience and updates it accordingly (Line 14).
Finally, the episode buffer is cleared in preparation for the next episode (Line 15).
The game terminates once the maximum number of training episodes is reached, after which the trading results are finalized.

\textbf{FL training guided by trading results:} After obtaining the trading results (MSPs' reward decisions $\bm{p}$ and resource allocation $(\bm{f},\bm{B})$), MUs and MSPs enter the FL phase. MUs follow the resource allocation scheme for local training and model uploading (Line 21). Then, the MSP is responsible for aggregating and updating the global model (Line 23). Finally, the updated model is provided for use by the AR engine (Line 24). This process iterates multiple times until the total FL time $T$ is reached. During real-time interactions, the MSP's AR service immersion and quality can be enhanced by utilizing the updated model.

\begin{table}[!t]
	\centering\caption{The Utilities of three reward approaches}
	\begin{tabular}{l|lll|l}
		\hline
		& MSP 1 & MSP 2 & MSP 3 & Total \\ \hline
		Optimal     & 2169.99     & 2564.56     & 3035.96    & 7770.51     \\ \hline
		MDDR  & 2169.11     & 2563.95     & 3034.94     & 7768.00     \\ \hline
		MAPPO & 2169.23     & 2563.96     & 3035.26     & 7768.45     \\ \hline
	\end{tabular}
	\label{DRL_OPTIMAL}
 \vspace{-0.1in}
\end{table}

\begin{figure*}
	\centering
	\subfigure[MSP 1] {
		\label{minist_MSP1}
		\includegraphics[scale=0.19]{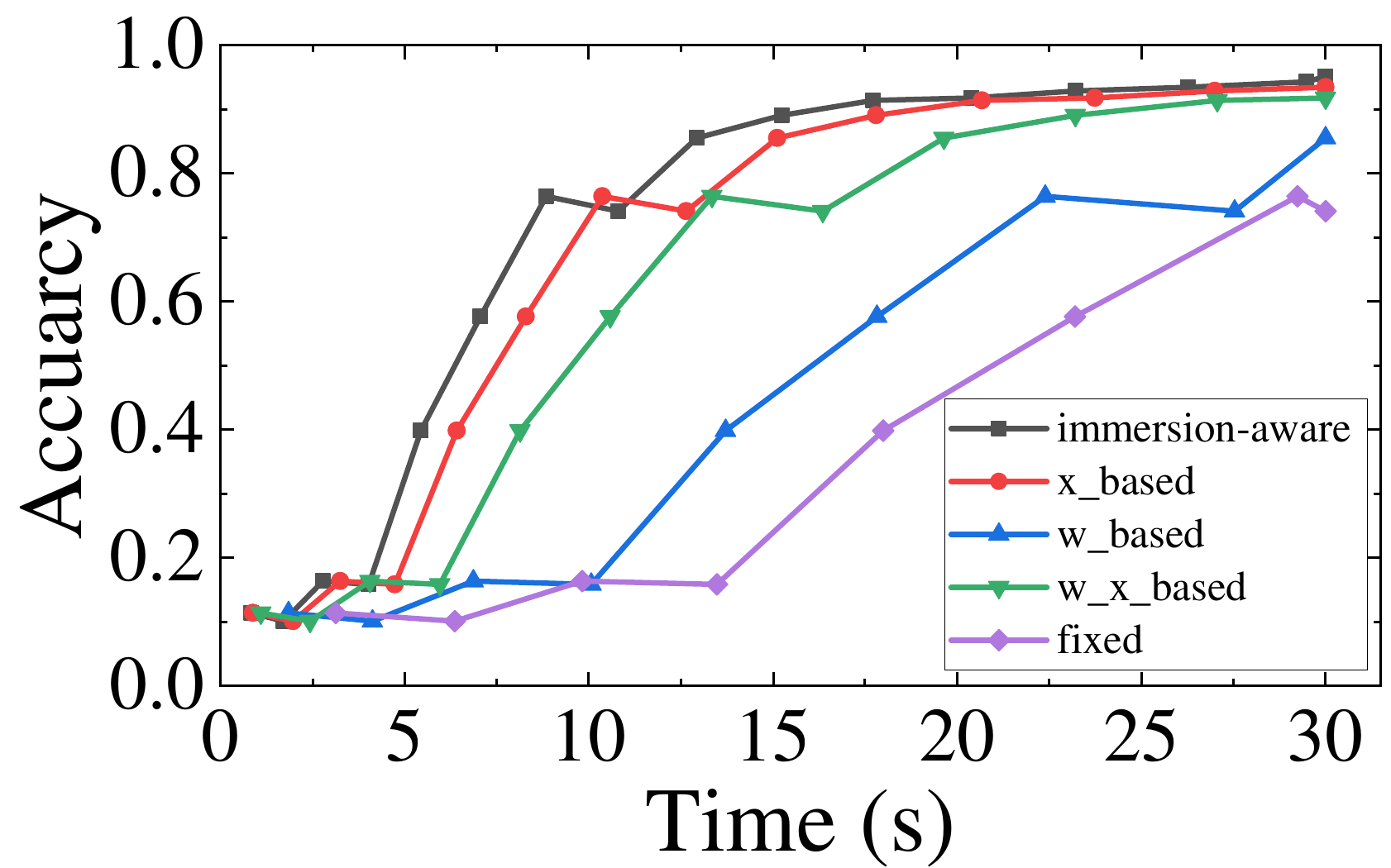}
	}
	\subfigure[MSP 2] {
		\label{minist_MSP2}
		\includegraphics[scale=0.19]{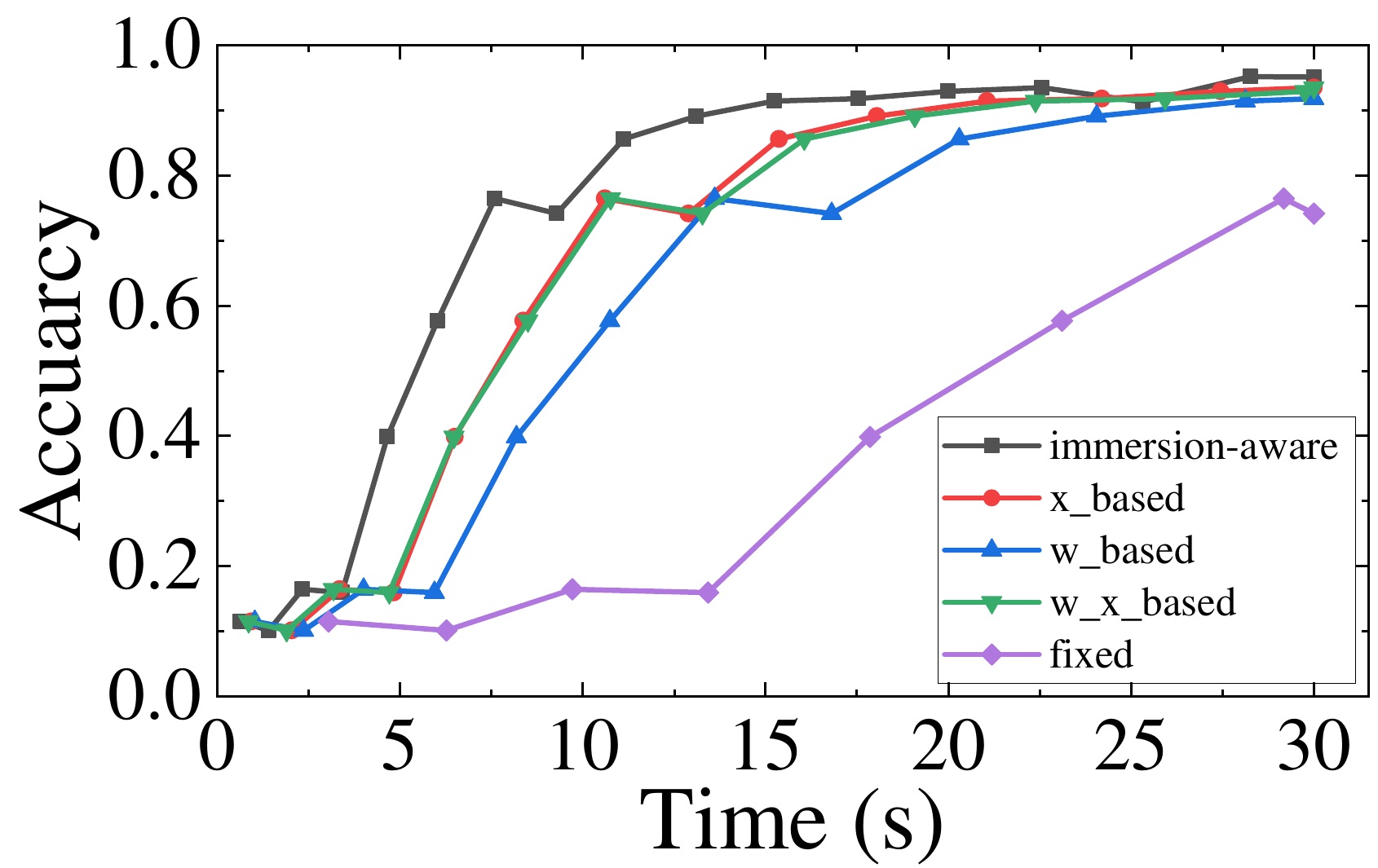}
	}
	\subfigure[MSP 3] {
		\label{minist_MSP3}
		\includegraphics[scale=0.19]{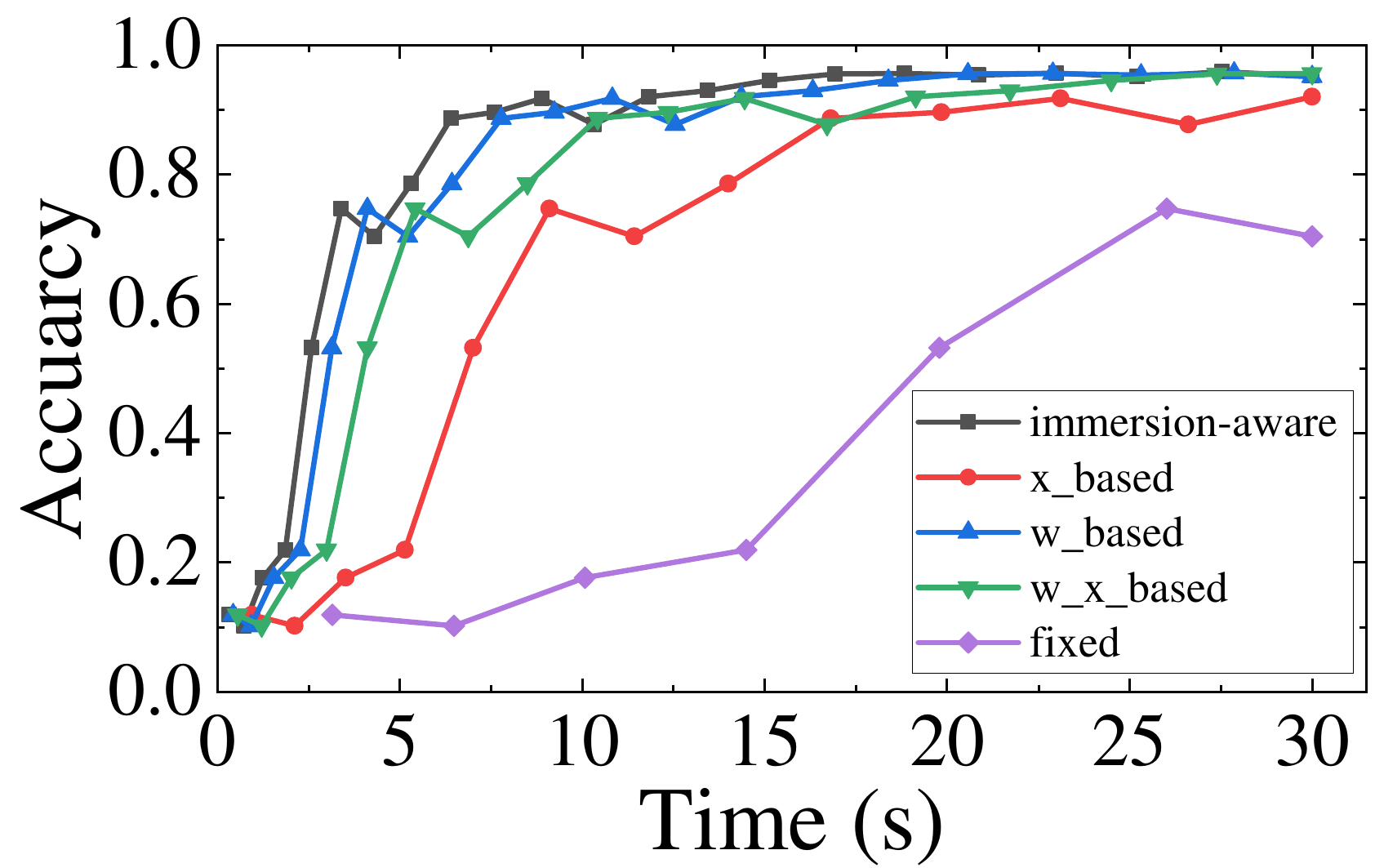}
	}
	\subfigure[The IoM for MSP 1] {
		\label{VOI_minist_MSP1}
		\includegraphics[scale=0.19]{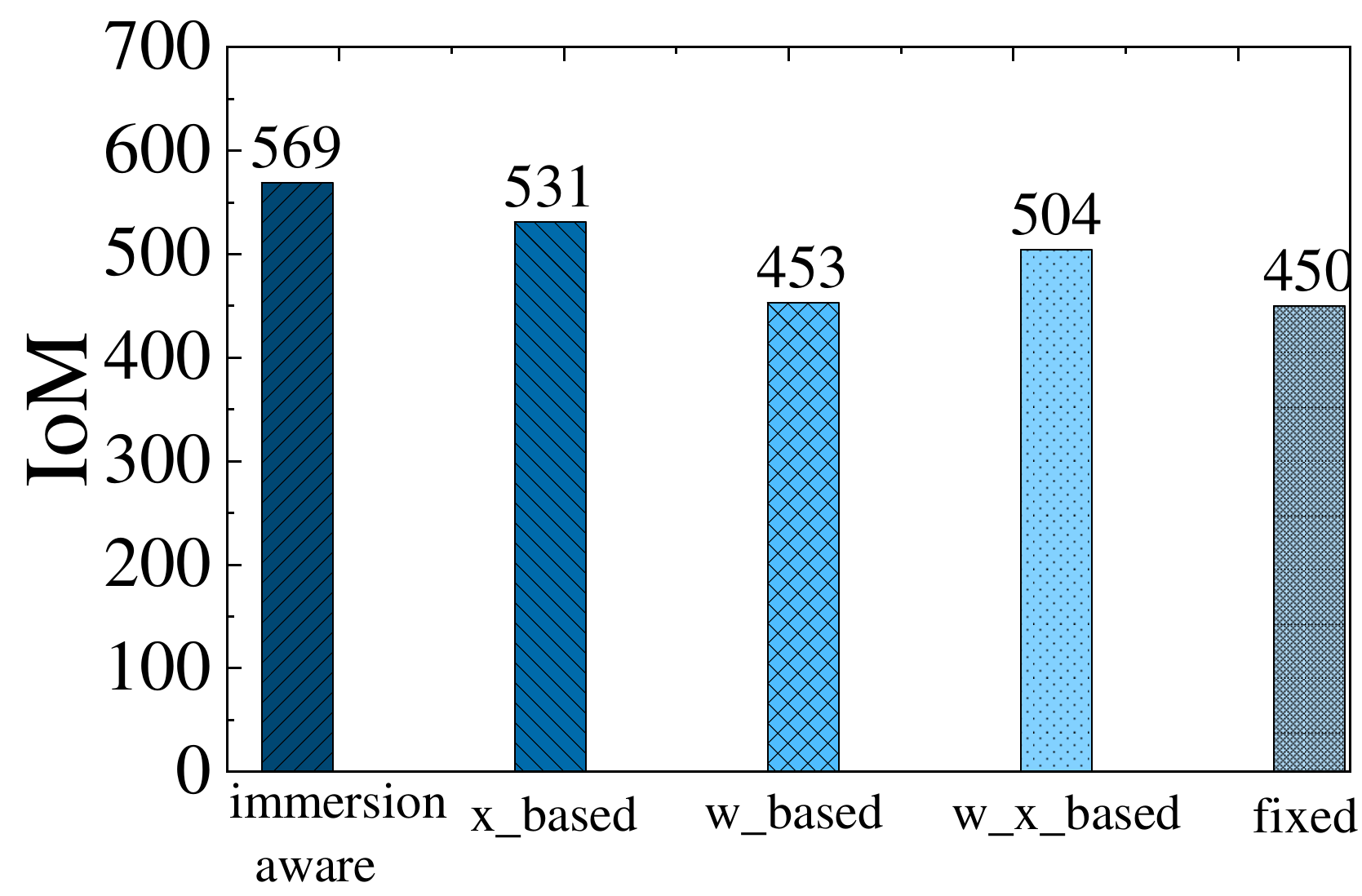}
	}
	\subfigure[The IoM for MSP 2] {
		\label{VOI_minist_MSP2}
		\includegraphics[scale=0.19]{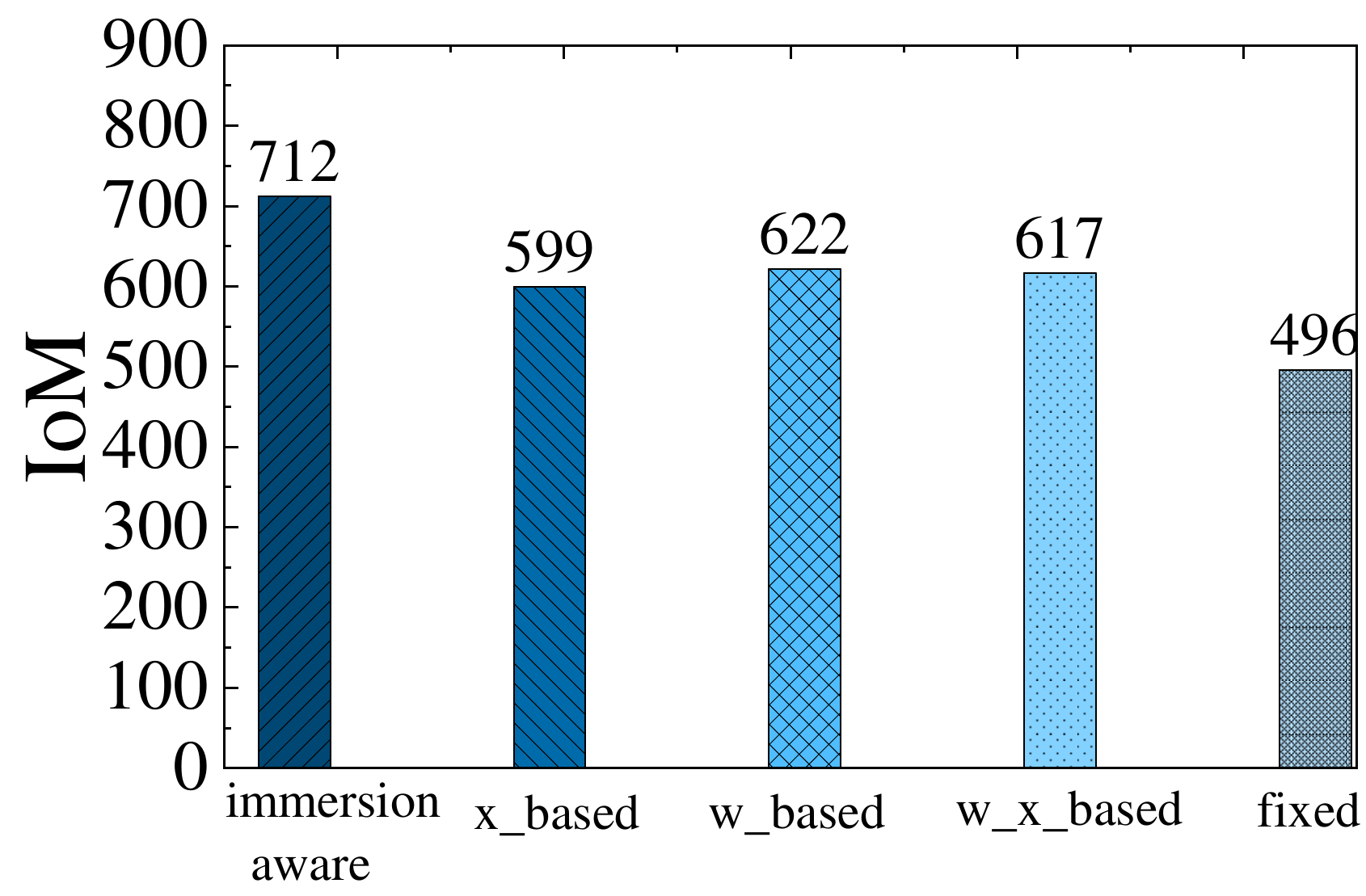}
	}
	\subfigure[The IoM for MSP 3] {
		\label{VOI_minist_MSP3}
		\includegraphics[scale=0.19]{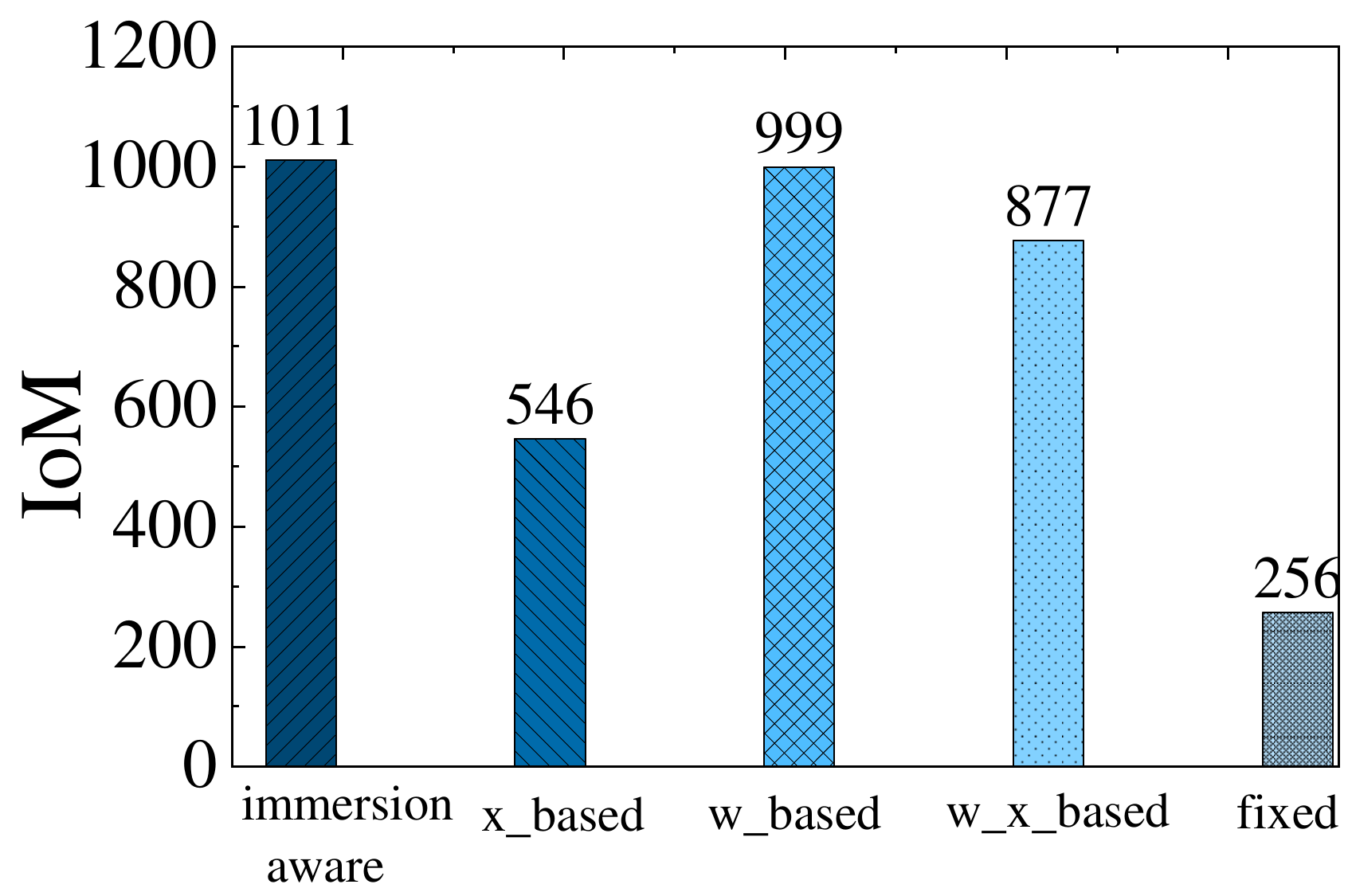}
	}
	\caption{Evaluation results for MNIST.}
	\label{minist}
\end{figure*}

\begin{figure*}
	\centering
	\subfigure[MSP 1] {
		\label{gtsrb_MSP1}
		\includegraphics[scale=0.19]{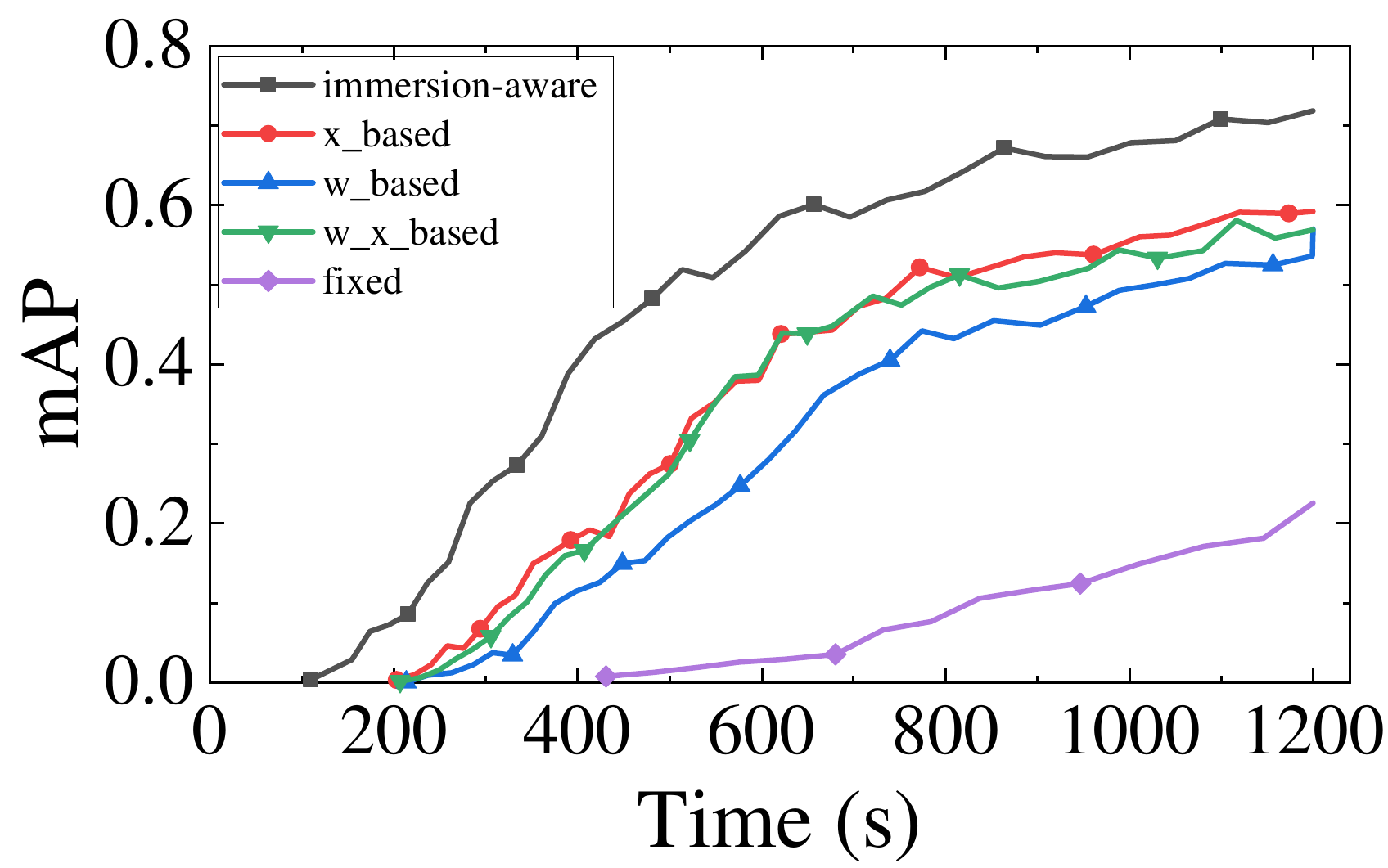}
	}
	\subfigure[MSP 2] {
		\label{gtsrb_MSP2}
		\includegraphics[scale=0.19]{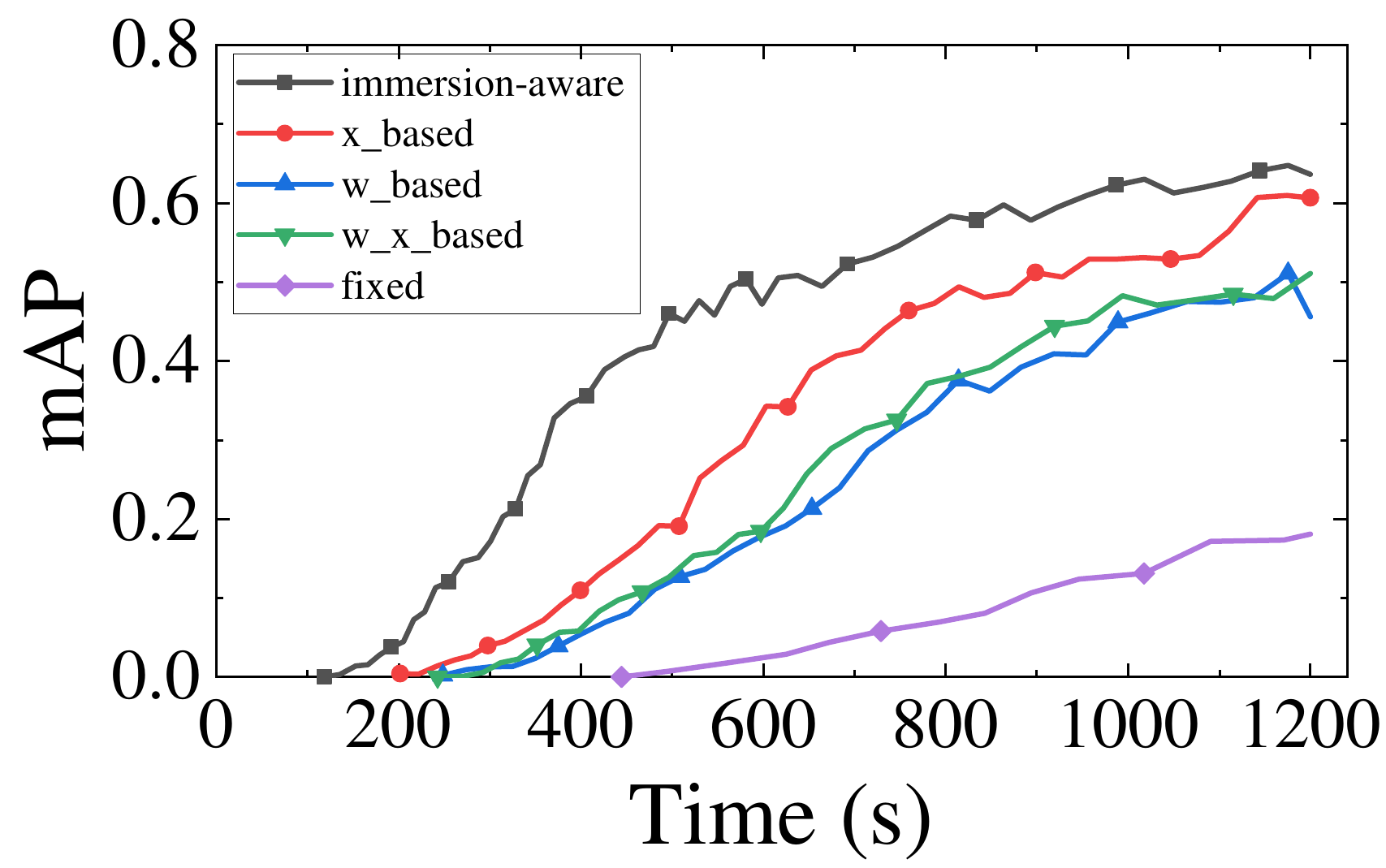}
	}
	\subfigure[MSP 3] {
		\label{gtsrb_MSP3}
		\includegraphics[scale=0.19]{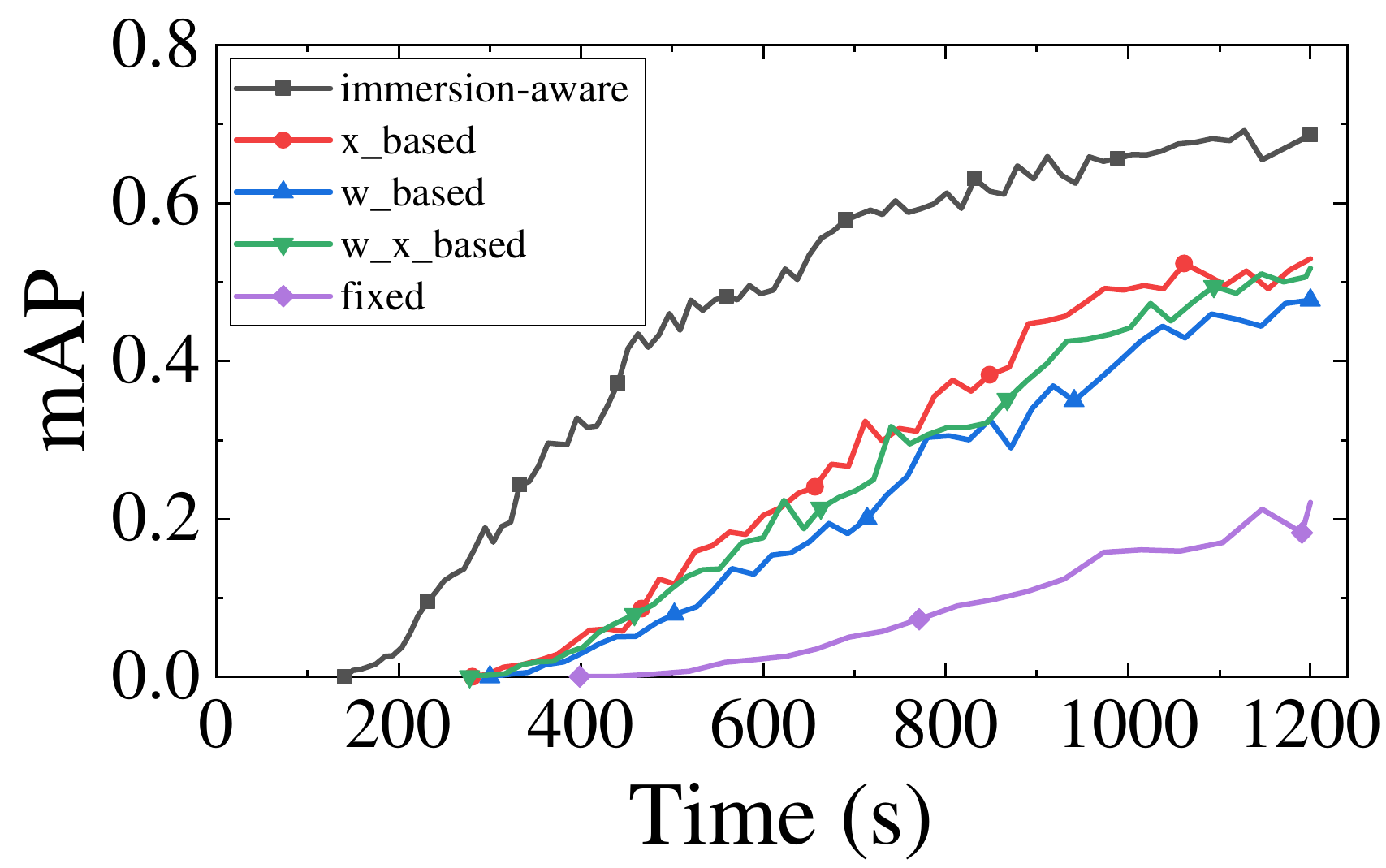}
	}
	\subfigure[The IoM for MSP 1] {
		\label{VOI_MC1_gtsrb}
		\includegraphics[scale=0.19]{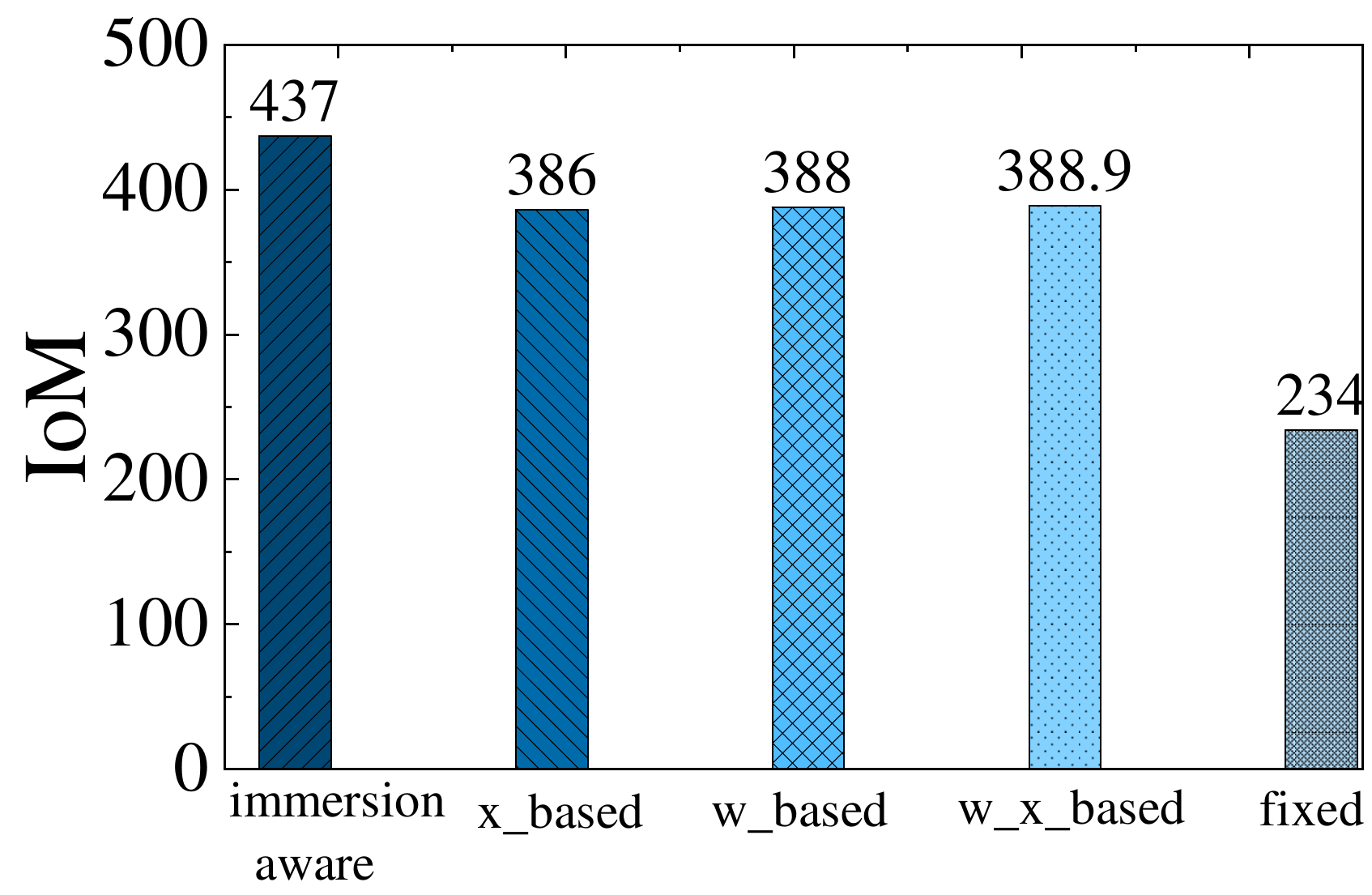}
	}
	\subfigure[The IoM for MSP 2] {
		\label{VOI_MC2_gtsrb}
		\includegraphics[scale=0.19]{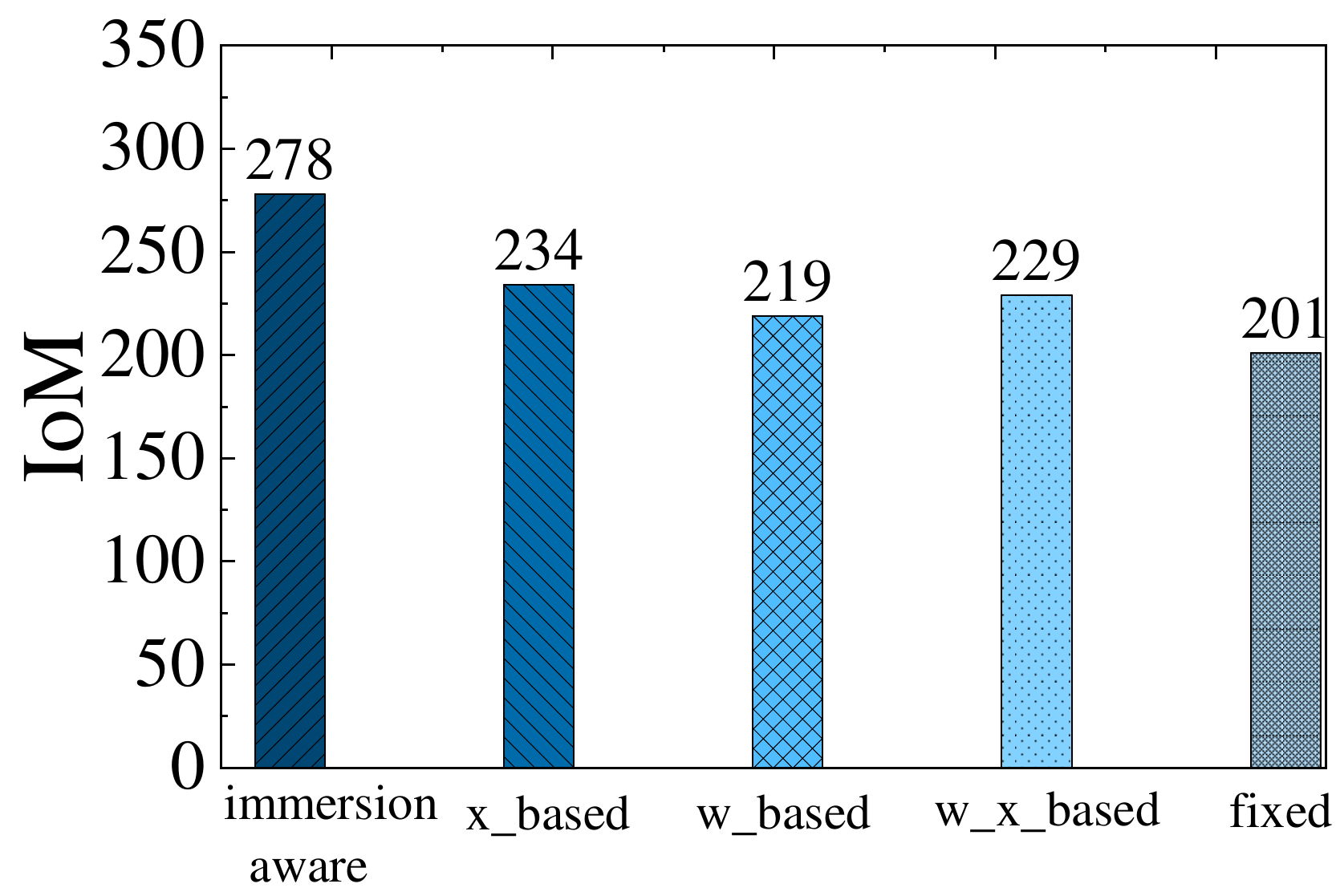}
	}
	\subfigure[The IoM for MSP 3] {
		\label{VOI_MC3_gtsrb}
		\includegraphics[scale=0.19]{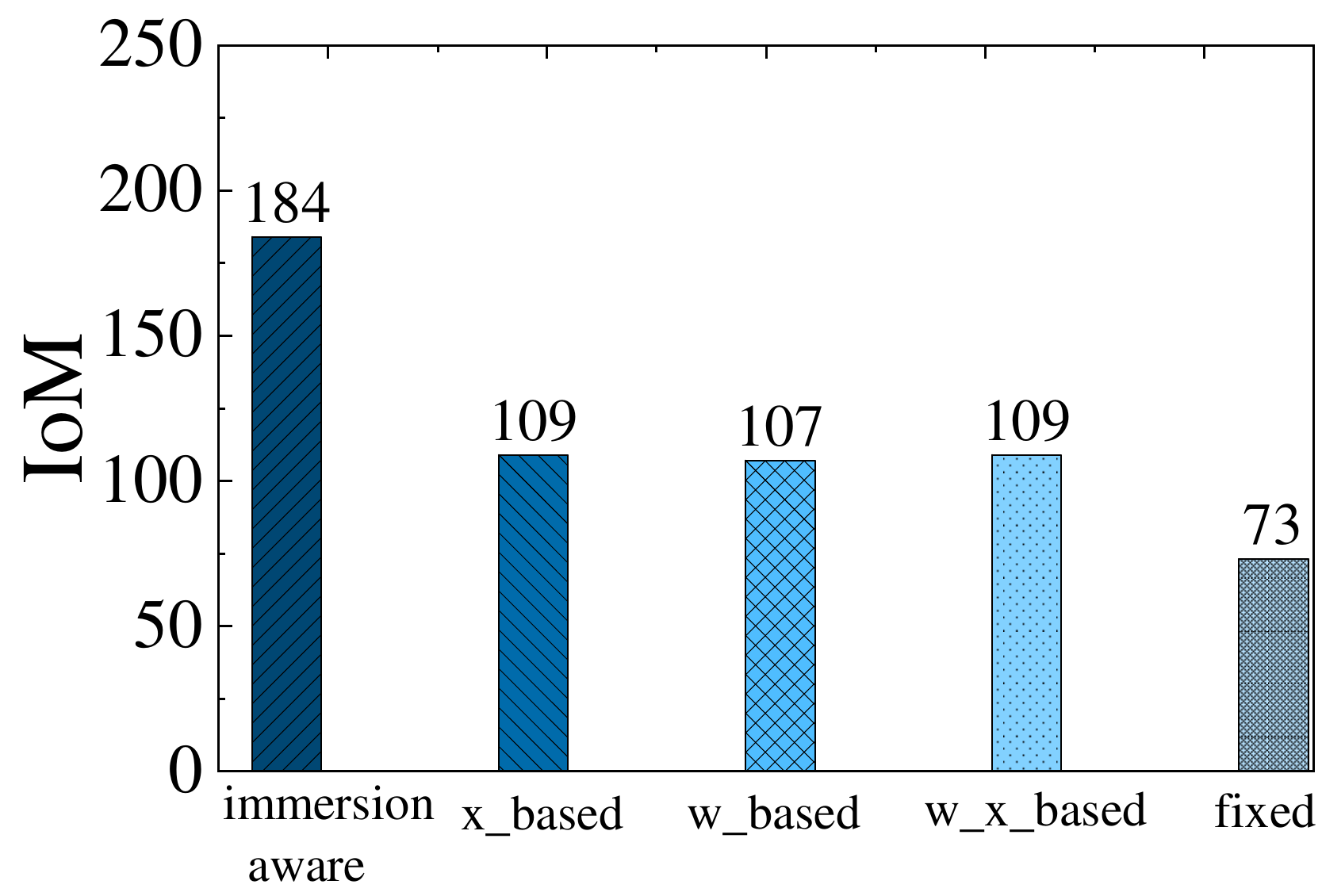}
	}
	\caption{Evaluation results for GTSRB.}
	\label{gtsrbgraph}
 \vspace{-0.1in}
\end{figure*}

\section{Performance Evaluation}\label{VII}
First, we verify the near-optimal performance of MDDR. Then, to evaluate the effectiveness of the proposed framework, we compare its performance with benchmark schemes in object detection and classification for AR services through FL.

\subsection{Simulation Settings}
We conduct experiments on the MNIST dataset with ResNet-18 and the GTSRB dataset with Faster R-CNN. 
The GTSRB was recorded during daytime driving on different types of roads in Germany for vehicle AR scenario testing~\cite{GTSRB1}.
We set the number of MUs as $M=5$ and the number of MSPs as $N=3$.
The configurations of the maximum computational and communication resource for each MU are randomly generated from the range $[3,5]$GHz and $[1, 4]$MHz~\cite{jovsilo2018selfish}. 
For local training of MUs, we adopt the SGD optimizer with $\text{momentum} =0.9$ and $\text{learning rate} = 0.001$.
For model aggregation, MSPs adopt the Federated Averaging (FedAvg) algorithm, applying equal weights to all participants' updates.  
The trading guidance time $T$ for MNIST and GTSRB are $30$s and $1,200$s, respectively. All baseline methods are configured with the same settings for fair comparison.

\subsection{Benchmark Schemes}
To our knowledge, there are no comparable solutions to the model trading problem in the vehicular metaverse. Therefore, under the uniform rewards of MSPs, we develop and extend four benchmark schemes for multi-MSP and multi-MU scenarios while meeting FL deadlines and resource constraints of devices.
\begin{itemize}{
		\item {\textbf{\textit{x\_based}}: Inspired by the incentive mechanism of the Stackelberg game for federated learning~\cite{Game},  the MU allocates its computational and communication resources based on the amount of data involved in local training.}
		
		\item {\textbf{\textit{w\_based}}: The MU adjusts the resource allocation according to the inference loss~\cite{zeng}, i.e., the potential value of local data in our work.}
		
		\item {\textbf{\textit{w\_x\_based}}: Combining the $\textbf{\textit{x\_based}}$ and $\textbf{\textit{w\_based}}$ designs, the MU allocates resources considering both the data size and the potential value of data.}
		
		\item {\textbf{\textit{fixed}}: Taking into account the differences in performance of the various devices that exist in reality, we randomly allocate a fixed set of computational and communication resources to reflect the heterogeneity of devices.}}
\end{itemize}

\subsection{Performance Evaluation of MDDR}
First, we consider a relatively static communication environment (fixed channel conditions) to show the effectiveness of MDDR compared to the optimal solution and the general multi-agent depth approximation policy optimization (MAPPO) approach. 
We trained our proposed MDDR and MAPPO until convergence, and the results are shown in Fig.~\ref{optimal} and Table~\ref{DRL_OPTIMAL}. We find that both MDDR and MAPPO achieve good results, but MAPPO slightly outperforms MDDR.
This is attributed to MAPPO's centrally controlled yet distributed execution architecture~\cite{10637735} that can utilize more information for decision making. However, this approach is not suitable for non-cooperation due to the need to share parameters among MSPs. 

For the fully distributed MDDR, it is more suitable for non-cooperative scenarios. The reason is that in MDDR, each agent has a self-contained actor-critic network, and information is not required to be shared. This suggests that the approach can achieve considerable utility while preserving privacy and saving communication costs significantly.
Then, we deploy the MDDR in dynamic communication environments where channel conditions change over time.
The trading behaviors of each MU and MSP are illustrated in Figs.~\ref{MU}, \ref{MSP1_price}, \ref{MSP2_price} and \ref{MSP3_price}, reflecting that the MDDR can converge rapidly and show good stability in dynamic network environments.

\subsection{Benchmark Comparison}
\begin{table*}[!t]
	\centering
	\caption{Time comparison of five schemes for achieving the same performance}
	\begin{threeparttable} 
	\begin{tabular}{cccccccccc}
		\hline
		\multirow{3}{*}{Schemes} & \multicolumn{3}{c}{MNIST}         & \multicolumn{6}{c}{GTSRB}                                         \\
		& \multicolumn{3}{c}{\underline{Training Time to reach 90\% Accuracy (s)}} & \multicolumn{3}{c}{\underline{Training Time to reach 50\% mAP (s)}}   & \multicolumn{3}{c}{\underline{Training Time to reach 70\% mAP@${50}$ (s)}} \\ 
		& MSP 1      & MSP 2     & MSP 3    & MSP 1    & MSP 2    & MSP 3    & MSP 1     & MSP 2    & MSP 3     \\ \hline
		immersion-aware          & 15.23     & 13.09    & 7.59    & 513.16  & 581.09  & 623.89  & 418.28   & 447.69  & 451.27   \\
		x\_based                 & 17.79     & 18.03    & 19.80   & 812.53  & 898.59  & 1061.50 & 596.57   & 679.91 & 890.47   \\
		w\_based                 & 30+       & 24.04    & 10.28   & 1064.83 & 1175.58 & 1200+   & 774.09   & 953.50  & 1013.89  \\
		w\_x\_based              & 23.20     & 19.08    & 12.34   & 901.61  & 1190    & 1194    & 596.35   & 848.68  & 932.73   \\
		fixed                    & 30+       & 30+      & 30+     & 1200+   & 1200+   & 1200+   & 1200+    & 1200+   & 1200+    \\ \hline
	\end{tabular}
\begin{tablenotes} 
\item Note: $30$+ or $1200$+ indicates that the scheme required more than $30$ seconds or $1200$ seconds, respectively, to achieve the target accuracy, exceeding the guideline time for the trading results. 
\end{tablenotes} 
\end{threeparttable} 
	\label{gtsrb}
 \vspace{-0.1in}
\end{table*}

Figs.~\ref{minist_MSP1}, \ref{minist_MSP2}, and \ref{minist_MSP3} show the performance improvement in classification accuracy over time, where the \textbf{\textit{immersion-aware}} outperforms the other schemes.
This is because \textbf{\textit{immersion-aware}} utilizes IoM to incentivize MUs to allocate computational and communication resources for providing local models to MSPs. The IoM metric not only evaluates the freshness and accuracy of the local model but also considers both the amount and potential value of raw data used for training. As a result, resource-constrained MUs are motivated to contribute more resources, providing valuable local models to MSPs more rapidly. 

In contrast, baseline schemes like  \textbf{\textit{x\_based}}, \textbf{\textit{w\_based}}, and \textbf{\textit{w\_x\_based}} struggle to offer adequate incentives for MUs to contribute their available resources due to uniform rewards. For example, if an MU has already reached its maximum allocatable resources at a given reward $p$, increasing $p$ further will not yield additional contributions. In this situation, raising the reward only increases the MSP’s cost without eliciting additional contributions, so its utility actually declines. Consequently, the uniform reward $p$ remains unchanged, preventing other MUs with more resources from contributing higher-value local models.

In addition, Figs.~\ref{minist_MSP1}, \ref{minist_MSP2}, and \ref{minist_MSP3} reveal that MSP 3 performs better than MSP 1 and MSP 2. For example, at the $10$th second, the accuracy of MSP 3 is about $88\%$, which is $14\%$ and $3\%$ higher than those of MSP 1 and MSP 2, respectively. This is because MSP 3 is associated with a higher potential value $\omega$, resulting in more substantial rewards being offered.
As a result, the MUs will allocate more computational and communication resources to provide local models for MSP 3 than for MSP 1 and MSP 2.
For the three different MSPs, some fluctuations are normal for model training (e.g., at time $10$s). 
Notably, the \textbf{\textit{fixed}} scheme performs the worst due to the limitations of synchronous FL, which is bottlenecked by the least capable MU.

Meanwhile, it is clear from Figs.~\ref{VOI_minist_MSP1},~\ref{VOI_minist_MSP2} and~\ref{VOI_minist_MSP3} that the effectiveness of the five schemes is influenced by IoM. With the higher IoM, the performance is improved faster, resulting in a better immersive experience. Although the IoM of \textbf{\textit{w\_based}} in Fig.~\ref{VOI_minist_MSP2} is higher than \textbf{\textit{x\_based}} and  \textbf{\textit{w\_x\_based}}, its performance improvement rate is comparatively slower.
This is because the IoM represents the aggregate value across all MUs, while the performance improvement rate is hampered by the worst MU.
Furthermore, from Figs.~\ref{minist_MSP2} and~\ref{minist_MSP3}, it can be observed that both \textbf{\textit{w\_x\_based}} and \textbf{\textit{w\_based}} perform better when compared to Fig.~\ref{minist_MSP1}. The main reason is the higher potential value $\omega$ associated with MSP 2 and MSP 3.

For object detection on GTSRB, 
Figs.~\ref{gtsrb_MSP1}, \ref{gtsrb_MSP2} and \ref{gtsrb_MSP3} show the mean average precision (mAP) and Table~\ref{gtsrb} gives the mAP@${50}$ (The value of mAP when the intersection over union (IOU) threshold is larger than $0.5$) results. Compared to MNIST, the performance enhancement for object detection is slower due to the complexity of the detection task. Specifically, it takes at least $513.16$ seconds to achieve a $50\%$ mAP using the \textbf{\textit{immersion-aware}} method. Our research introduces the flexibility to adjust decision guidance time, making it adaptable to various task types. Moreover, Figs.~\ref{VOI_MC1_gtsrb},~\ref{VOI_MC2_gtsrb},~\ref{VOI_MC3_gtsrb} and Table~\ref{gtsrb} demonstrate that MSP 1 with greater IoM takes less time to achieve the same mAP, which indicates that IoM works well to capture immersion. For instance, MSP 1 has the highest IoM, and the time that it takes to reach $50\%$ mAP is $513.16$s, which is $67.93$s and $110.73$s less than those of MSP 2 and MSP 3, respectively.

The comparison results validate the advantages of the proposed trading framework, which outperforms other schemes by improving IoM by $38.3\%$ and $37.2\%$ and reducing the training time to reach the target accuracy by $43.5\%$ and $49.8\%$, on average, for the MNIST and GTSRB datasets, respectively.

\section{Conclusion}\label{VIII}
To facilitate the development of vehicular metaverse services, we propose an immersion-aware model trading framework, incorporating FL to incentivize MUs to contribute local models for AR services such as object detection and classification. Specifically, we construct an EPEC problem with MSPs as leaders and MUs as followers to achieve an equilibrium among their interests. A new metric called ``IoM'' is designed to comprehensively evaluate the enhancement brought by the local models of MUs for AR services.
Furthermore, considering the competitive relationship among MSPs and the dynamic network environment, we develop a fully distributed MDDR approach to obtain the reward decisions of MSPs. Extensive simulations on AR-related vehicle and MNIST datasets demonstrate that the proposed framework enables more efficient and immersive AR services in the vehicular metaverse.

\appendices

\section{Proof of Proposition 1}
\begin{proof}
	First, we transform \textbf{Problem 1} into an unconstrained optimization problem using the Lagrangian dual method~\cite{1035044, 10870187}. The Lagrangian function associated with MU $m$ is formulated as
	\begin{equation}
	\begin{aligned}
&\ L_{m}\left(\bm{f}_{m},\bm{B}_{m},\bm{p}_{m},\lambda_{m},\beta_{m}, \delta _{mn}, \gamma_{m} \right)\\ =&\ \sum_{n} p_{mn}V_{mn} -\sum_{n}\left[c_{m}^{f}f_{mn}\log(1/\theta_{m})+ c_{m}^{B}B_{mn}\right] \\&-\lambda_{m}(\sum_{n}f_{mn}-f_{m}^{max})-\beta_{m}(\sum_{n}B_{mn}-B_{m}^{max})\\&-\sum_{n}\delta_{mn}[\log(1/\theta _{m})\frac{ x_{m}\tau_{n}}{f_{mn}}+\frac{b_{mn}}{B_{mn}\log_{2}(1+\varsigma_{mn})}- \frac{1}{2}\tau_{n}]\\&-\gamma_{m}[ S_{m}-T^{req} ( f_{m}^{max}-\sum_{n}f_{mn})],
	\end{aligned}
	\end{equation} where ($\lambda _{m}\geq 0$ and $\beta _{m}\geq 0$), ($ \delta_{mn}\geq 0$), and ($\gamma_{m}\geq0$) are the Lagrangian dual variables
	corresponding to constraints C1, C2, and C3, respectively. 
    Then, the first-order derivative of $L_{m}$ with respect to $f_{mn}$ and $B_{mn}$ can be derived as
	\begin{equation}\label{ff}
	\begin{aligned}
	\frac{\partial L_{m}}{\partial f_{mn}}= &\ p_{mn} I_{mn}\frac{x_{m}\tau_{n}\log(1/\theta_{m})}{f_{mn}^{2}}-c_{m}^{f}\log(1/\theta_{m})\\&-\lambda _{m}+\delta _{mn}\log(1/\theta_{m})\frac{ x_{m}\tau_{n}}{f_{mn}^{2}}-\gamma_{m}T^{req},
	\end{aligned}
	\end{equation} and
	\begin{equation}
	\begin{aligned}
	\frac{\partial L_{m}}{\partial B_{mn}}= &\ p_{mn} I_{mn}\frac{b_{mn}}{B_{mn}^{2}\log_{2}(1+\varsigma_{mn})}-c_{m}^{B}-\beta_{m}\\&+\delta _{mn}\frac{b_{mn}}{B_{mn}^{2}\log_{2}(1+\varsigma_{mn})},
	\end{aligned}
	\end{equation}
respectively.
From (\ref{ff}) and the constraints in (\ref{pro1c}), the Karush-Kuhn-Tucker (KKT)~\cite{BoydVandenberghe2004} conditions are given by
\begin{equation}
\begin{aligned}
&\frac{\partial L_{m}}{\partial f_{mn}}=0, \  \lambda_{m}(\sum_{n}f_{mn}-f_{m}^{max})=0,\\
&\delta_{mn}[\log(1/\theta _{m})\frac{x_{m}\tau_{n}}{f_{mn}}+\frac{b_{mn}}{B_{mn}\log_{2}(1+\varsigma_{mn})}-\frac{1}{2}\tau_{n}]=0,\\
&\gamma_{m} [ S_{m}-T^{req} ( f_{m}^{max}-\sum_{n}f_{mn})]=0,\\
&\lambda_{m}, \delta_{mn}, \gamma_{m}\geq  0,\ C1,\ C2,\ C3. 
\end{aligned}
\end{equation}
Since $ S_{m}<T^{req} \left( f_{m}^{max}-\sum_{n}f_{mn} \right)$ and $\sum_{n}f_{mn}<f_{m}^{max}$, we can obtain $\lambda_{m}=0$ and $\gamma_{m}=0$. Furthermore, $\frac{\partial L_{m}}{\partial f_{mn}}=0$ can be converted to
\begin{equation}\label{fxi=0}
(p_{mn} I_{mn}+\delta_{mn})\frac{x_{m}\tau_{n}}{f_{mn}^{2}}-c_{m}^{f}=0.
\end{equation}
Similarly, the KKT conditions for $B_{mn}$ are obtained as
\begin{equation}
\begin{aligned}
&\frac{\partial L_{m}}{\partial B_{mn}}=0, \  
\beta_{m}(\sum_{n}B_{mn}-B_{m}^{max})=0,\\
&\delta_{mn}[\log(1/\theta _{m})\frac{ x_{m}\tau_{n}}{f_{mn}}+\frac{b_{mn}}{B_{mn}\log_{2}(1+\varsigma_{mn})}-\frac{1}{2}\tau_{n}]=0,\\
&\beta_{m}, \delta_{mn}\geq  0,\ C1,\ C2. 
\end{aligned}
\end{equation}
Since $\sum_{n}B_{mn}<B_{m}^{max}$, we can obtain $\beta_{m}=0$. Furthermore, $\frac{\partial L_{m}}{\partial B_{mn}}=0$ can be converted to
\begin{equation}\label{Bxi=0}
(p_{mn} I_{mn}+\delta_{mn})\frac{b_{mn}}{B_{mn}^{2}\log_{2}(1+\varsigma_{mn})}-c_{m}^{B}=0.
\end{equation}

Then, based on the value of $\delta_{mn}$, the optimal allocation of computational and communication resources exists in the following two cases:
\begin{itemize}
    \item {($\delta_{mn}=0$): According to (\ref{fxi=0}) and (\ref{Bxi=0}), we have
\begin{equation}
\begin{aligned}
    f_{mn}= \sqrt{\frac{p_{mn} I_{mn} x_{m}\tau_{n}}{ c_{m}^f}}, \ 
B_{mn}= \sqrt{\frac{p_{mn} I_{mn}b_{mn}}{ c_{m}^B \log_{2}(1+\varsigma_{mn})}}.
\end{aligned}
\end{equation} }
\item { ($\delta_{mn}>0$): Substituting $f_{mn}$ and $B_{mn}$ into $\delta_{mn}\left[\log(1/\theta _{m})\frac{ x_{m}\tau_{n}}{f_{mn}}+\frac{b_{mn}}{B_{mn}\log_{2}(1+\varsigma_{mn})}-\frac{1}{2}\tau_{n}\right]=0$, we can get
\begin{equation}
\begin{aligned}
&\delta_{mn}=F^{2} -p_{mn}I_{mn}, \\ &F=2\log(1/\theta_{m}) \sqrt{\frac{x_{m}c_{m}^{f}}{\tau_{n}} } +\frac{2\sqrt{b_{mn}c_{m}^{B}}}{\tau_{n}\sqrt{\log_{2}(1+\varsigma_{mn})} }.
\end{aligned}
\end{equation}

Next, bring $\delta_{mn}$ into $f_{mn}$ and $B_{mn}$, we have
\begin{equation}
\begin{aligned}
f_{mn}= F\sqrt{\frac{x_{m}\tau_{n}}{ c_{m}^f}}, \ B_{mn}= F\sqrt{\frac{b_{mn}}{ c_{m}^B \log_{2}(1+\varsigma_{mn})}}.
\end{aligned}
\end{equation}}
\end{itemize}
 \end{proof}

\section{Proof of Theorem 2}
\begin{proof}
First, we need to compute $\frac{\partial f_{mn}}{\partial p_{mn}}$, $\frac{\partial^{2} f_{mn}}{\partial p_{mn}^{2}}$, $\frac{\partial B_{mn}}{\partial p_{mn}}$, and $\frac{\partial^{2} B_{mn}}{\partial p_{mn}^{2}}$ to obtain the values of $V_{mn}'$ and $V_{mn}''$.
From (\ref{fxi=0}), we can obtain $\frac{\partial f_{mn}}{\partial p_{mn}}$ and $\frac{\partial^{2} f_{mn}}{\partial p_{mn}^{2}}$,
the steps of which are shown as follows. The first-order derivative of $\frac{\partial L_{m}}{\partial f_{mn}}=0$ with respect to $p_{mn}$ is expressed as $I_{mn}\frac{x_{m}\tau_{n}}{f_{mn}^{2}}-2(p_{mn} I_{mn}+\delta _{mn})\frac{x_{m}\tau_{n}}{f_{mn}^{3}}\cdot \frac{\partial f_{mn}}{\partial p_{mn}}=0$. Consequently, we can obtain
\begin{equation}\label{f1}
	\frac{\partial f_{mn}}{\partial p_{mn}}=\frac{I_{mn}f_{mn}}{2(p_{mn}I_{mn}+\delta_{mn})}>0,
\end{equation}
the second-order derivative of $\frac{\partial L_{m}}{\partial f_{mn}}=0$ with respect to $p_{mn}$
is expressed as 
\begin{equation}\label{f2}
\begin{aligned}
\frac{\partial^{2} f_{mn}}{\partial p_{mn}^{2}}=\frac{I_{mn}\frac{\partial f_{mn}}{\partial p_{mn}}(p_{mn}I_{mn}+\delta_{mn})-I_{mn}^{2}f_{mn}}{2(p_{mn}I_{mn}+\delta_{mn})^{2}}.
\end{aligned}
\end{equation}
By substituting (\ref{f1}) into (\ref{f2}), we have
\begin{equation}\label{f3}
\frac{\partial^{2} f_{mn}}{\partial p_{mn}^{2}}=\frac{-I_{mn}^{2}f_{mn}}{4(p_{mn}I_{mn}+\delta_{mn})^{2}}<0.
\end{equation}

Likewise, we derive $\frac{\partial B_{mn}}{\partial p_{mn}}$ and $\frac{\partial^{2} B_{mn}}{\partial p_{mn}^{2}}$ with
the following steps based on (\ref{Bxi=0}). The first-order derivative of $\frac{\partial L_{m}}{\partial B_{mn}}=0$
with respect to $p_{mn}$ is expressed as $I_{mn}\frac{b_{mn}}{B_{mn}^{2}\log_{2}(1+\varsigma_{mn})}-2(p_{mn} I_{mn}+\delta _{mn})\frac{b_{mn}}{B_{mn}^{3}\log_{2}(1+\varsigma_{mn})}\cdot \frac{\partial B_{mn}}{\partial p_{mn}}=0$. Accordingly, we have
\begin{equation}\label{B1}
\frac{\partial B_{mn}}{\partial p_{mn}}=\frac{I_{mn}B_{mn}}{2(p_{mn}I_{mn}+\delta_{mn})}>0.
\end{equation}
Similarly, we obtain the second-order derivative of $\frac{\partial L_{m}}{\partial B_{mn}}=0$
with respect to $p_{mn}$ as 
\begin{equation}\label{B2}
\begin{aligned}
\frac{\partial^{2} B_{mn}}{\partial p_{mn}^{2}}=\frac{I_{mn}\frac{\partial B_{mn}}{\partial p_{mn}}(p_{mn}I_{mn}+\delta_{mn})-I_{mn}^{2}B_{mn}}{2(p_{mn}I_{mn}+\delta_{mn})^{2}}.
\end{aligned}
\end{equation} 
By substituting (\ref{B1}) into (\ref{B2}), we have
\begin{equation}\label{B3}
\frac{\partial^{2} B_{mn}}{\partial p_{mn}^{2}}=\frac{-I_{mn}^{2}B_{mn}}{4(p_{mn}I_{mn}+\delta_{mn})^{2}}<0.
\end{equation}

\begin{small}
    \begin{figure*}[!t]
	\normalsize
	\begin{equation}\label{value}
	\begin{aligned}
	V_{mn}'=\frac{\partial V_{mn}}{\partial p_{mn}}=I_{mn}&\left ( \frac{ x_{m}\tau_{n}\log(1/\theta_{m})}{f_{mn}^{2}}\cdot \frac{\partial f_{mn}}{\partial p_{mn}} +\frac{b_{mn}}{B_{mn}^{2}\log_{2}(1+\varsigma_{mn})}\cdot\frac{\partial B_{mn}}{\partial p_{mn}} \right )>0,\\ V_{mn}''= \frac{\partial^{2} V_{mn}}{\partial p_{mn}^{2}}=I_{mn}&\left [ \frac{-2x_{m}\tau_{n}\log(1/\theta_{m})}{f_{mn}^{3}}\cdot (\frac{\partial f_{mn}}{\partial p_{mn}})^{2}+ \frac{ x_{m}\tau_{n}\log(1/\theta_{m})}{f_{mn}^{2}}\cdot \frac{\partial^{2} f_{mn}}{\partial p_{mn}^{2}}\right.\\
	&\left. - \frac{2b_{mn}}{B_{mn}^{3}\log_{2}(1+\varsigma_{mn})}\cdot (\frac{\partial B_{mn}}{\partial p_{mn}})^{2}+\frac{b_{mn}}{B_{mn}^{2}\log_{2}(1+\varsigma_{mn})}\cdot\frac{\partial^{2} B_{mn}}{\partial p_{mn}^{2}} \right ]<0.
	\end{aligned}
	\end{equation}
\end{figure*}
\end{small}
\begin{small}
    \begin{figure*}[!t]
	\normalsize
	\begin{equation}
	\begin{aligned}
	\label{result}
  & \frac{\partial^{2} \Psi_{n}}{\partial p_{mn}^{2}}=-\Bigg[
\frac{aI_{mn}^3p_{mn}+4aI_{mn}^2\delta_{mn}}{4f_{mn}(p_{mn}I_{mn}+\delta_{mn})^2}
+ \frac{bI_{mn}^3p_{mn}+4bI_{mn}^2\delta_{mn}}{4B_{mn}(p_{mn}I_{mn}+\delta_{mn})^2}  + \frac{ \mu_{n}I_{mn}^2\left ( \frac{aI_{mn}}{2f_{mn}(p_{mn}I_{mn}+\delta_{mn})} +\frac{bI_{mn}}{2B_{mn}(p_{mn}I_{mn}+\delta_{mn})}\right )^2}{\left ( 1+\sum_{m} V_{mn} \right )^2 }  \\
& \qquad \qquad \qquad  +  \frac{ \mu_{n} I_{mn} \left (\frac{3aI_{mn}^2}{4f_{mn}(p_{mn}I_{mn}+\delta_{mn})^2}+\frac{3bI_{mn}^2}{4B_{mn}(p_{mn}I_{mn}+\delta_{mn})^2} \right )}{\left ( 1+\sum_{m} V_{mn} \right ) }\Bigg ]   <0, \\
&\text{where}\, a=x_{m}\tau_{n}\log(1/\theta_{m}), \ b=\frac{b_{mn}}{\log_{2}(1+\varsigma_{mn})}.
	\end{aligned}
	\end{equation}
	\hrulefill 
\end{figure*}
\end{small}

Based on (\ref{f1}), (\ref{f3}), (\ref{B1}), and (\ref{B3}), we can easily infer that $V_{mn}'>0$ and $V_{mn}''<0$, as shown in (\ref{value}). Then, we can obtain $\frac{\partial^{2} \Psi_{n}}{\partial p_{mn}\partial p_{m'n}}<0$. For  $\frac{\partial^{2} \Psi_{n}}{\partial p_{mn}^{2}}$, we substitute the specific expressions for $V_{mn}'$ and $V_{mn}''$ into (\ref{last-VV}) and obtain the following result as shown in (\ref{result}).
Next, by randomly selecting a vector $\bm{h}\in \mathbb{R}^{M\times 1}$ with elements not all $0$, we obtain
\begin{align}
&\ \bm{h}^\top (\bm{\Lambda}_n + \bm{H}_n) \bm{h} \notag \\ = &\underbrace{\sum_{m=1}^M \frac{\partial^2 \Psi_n}{\partial p_{mn}^2} (h^m)^2}_{\text{diagonal part}} + \underbrace{\sum_{m \neq m'} h^m \frac{\partial^2 \Psi_n}{\partial p_{mn} \partial p_{{m'}n}} h^{m'}}_{\text{off-diagonal part}}.
\end{align}Expanding the expressions and letting $S = \sum_{m=1} V_{mn}$, we have
\begin{align}
&\ \quad \bm{h}^\top (\bm{\Lambda}_n + \bm{H}_n) \bm{h} 
\notag \\ &= \sum_{m \neq {m'}} h^m \left( -\mu_n \frac{V_{mn}' V_{{m'}n}'}{(1+S)^2} \right) h^{m'} \notag \\&+
\sum_{m=1}^M \left[ \mu_n \frac{V_{mn}''(1+S) - (V_{mn}')^2}{(1+S)^2} 
     - 2V_{mn}' - p_{mn} V_{mn}'' \right] (h^m)^2.
\end{align}
Specially, the off-diagonal part is transformed using the identity
\begin{align}
&\ \sum_{m \neq m'} h^m V_{mn}' V_{{m'}n}' h^{m'} \notag \\
= &\left( \sum_{m=1}^M h^m V_{mn}' \right)^2 - \sum_{m=1}^M (h^m)^2 (V_{mn}')^2,
\end{align}
thus,
\begin{align}
&\sum_{m \neq {m'}} h^m \left( -\mu_n \frac{V_{mn}' V_{{m'}n}'}{(1+S)^2} \right) h^{m'} \notag 
\\= &-\frac{\mu_n}{(1+S)^2} \left[ \left( \sum_{m=1}^M h^m V_{mn}' \right)^2 - \sum_{m=1}^M (h^m)^2 (V_{mn}')^2 \right].
\end{align}
Then, combining both parts, we have
\begin{align}
&\ \bm{h}^\top (\bm{\Lambda}_n + \bm{H}_n) \bm{h} \notag 
\\= &\sum_{m=1}^M \left[ \mu_n \frac{V_{mn}''(1+S) - (V_{mn}')^2}{(1+S)^2} - 2V_{mn}' - p_{mn}V_{mn}'' \right] (h^m)^2 \nonumber \\
&\quad - \frac{\mu_n}{(1+S)^2} \left[ \left( \sum_{m=1}^M h^m V_{mn}' \right)^2 - \sum_{m=1}^M (h^m)^2 (V_{mn}')^2 \right].
\end{align}
Moreover, the expression can be simplified to
\begin{align}
&\ \bm{h}^\top (\bm{\Lambda}_n + \bm{H}_n) \bm{h} \notag \\
= &-\frac{\mu_n}{(1+S)^2} \left( \sum_{m=1}^M h^m V_{mn}' \right)^2 \nonumber \\
&+ \sum_{m=1}^M (h^m)^2 \left( \mu_n \frac{V_{mn}''}{1+S} - 2V_{mn}' - p_{mn}V_{mn}'' \right).
\end{align}
Finally, we analyze the sign of this expression:
\begin{enumerate}
    \item Since $\mu_n > 0$ and $(1+S)^2 > 0$, the first term is non-positive.
    
    \item Plugging (\ref{value}) into the second term yields 
$( \mu_n \frac{V_{mn}''}{1+S} - 2V_{mn}' - p_{mn}V_{mn}'')<0$, as shown in (\ref{second-term}).

\begin{small}
    \begin{figure*}[!t]
	\normalsize
	\begin{equation}
	\begin{aligned}
	\label{second-term}
  & \mu_n \frac{V_{mn}''}{1+S} - 2V_{mn}' - p_{mn}V_{mn}''=-\Bigg[
\frac{aI_{mn}^3p_{mn}+4aI_{mn}^2\delta_{mn}}{4f_{mn}(p_{mn}I_{mn}+\delta_{mn})^2}
+  \frac{ \mu_{n} I_{mn} \left (\frac{3aI_{mn}^2}{4f_{mn}(p_{mn}I_{mn}+\delta_{mn})^2}+\frac{3bI_{mn}^2}{4B_{mn}(p_{mn}I_{mn}+\delta_{mn})^2} \right )}{\left ( 1+S \right ) }\Bigg ]   <0, \\
&\text{where}\, a=x_{m}\tau_{n}\log(1/\theta_{m}), \ b=\frac{b_{mn}}{\log_{2}(1+\varsigma_{mn})}.
	\end{aligned}
	\end{equation}
	\hrulefill 
\end{figure*}
\end{small}
\end{enumerate}

Based on the above, we can have $\bm{h}^{T}(\bm{\Lambda}_{n}+\bm{H_{n}})\bm{h}<0$, indicating that the utility function $\Psi_{n}$ is strictly concave. According to~\cite{10415185}, there exists a unique Nash equilibrium in the multi-MSP rewarding game $\Omega$. 
\end{proof}

\bibliographystyle{IEEEtran}
\bibliography{meta}

\begin{IEEEbiography}[{\includegraphics[width=1in,height=1.25in,clip,keepaspectratio]{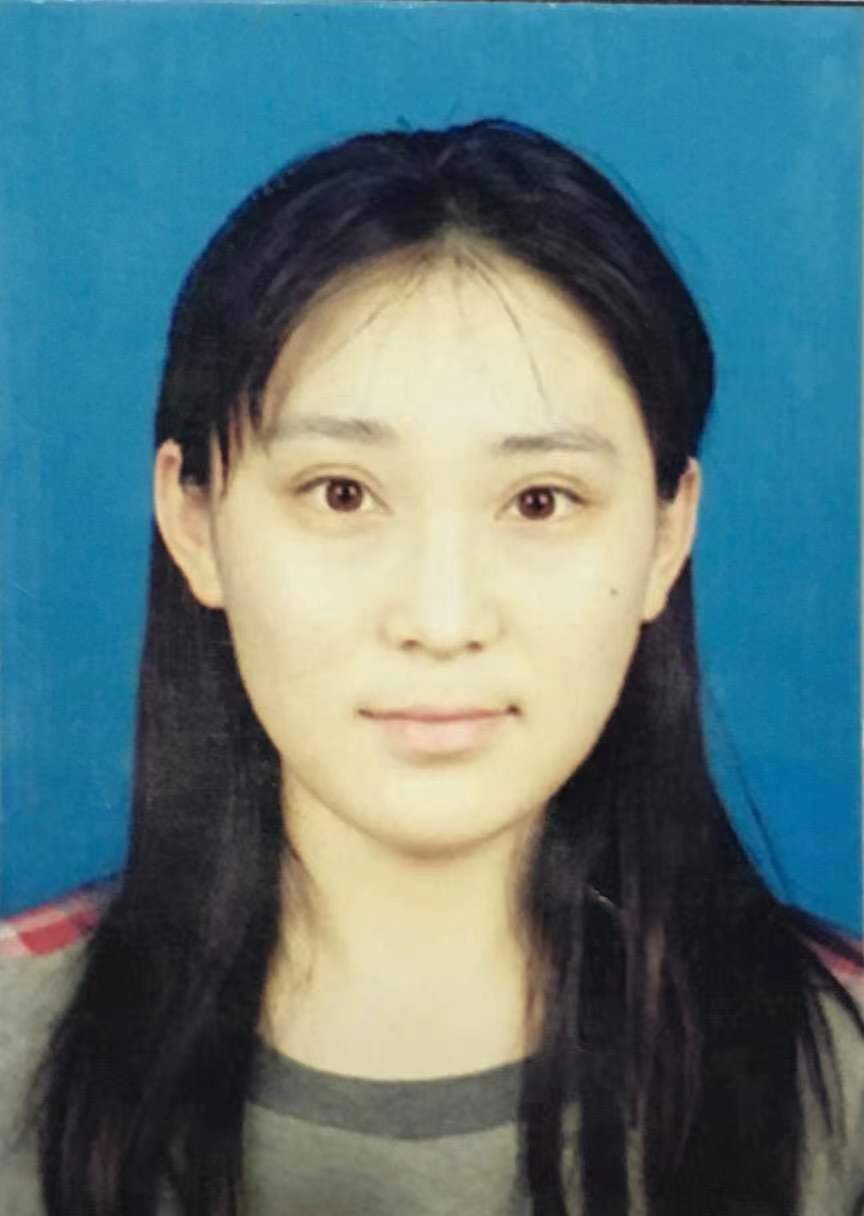}}]{Hongjia Wu} received the Ph.D. degree in computer science from the National University of Defense Technology, China, in 2023. She was a visiting student at the Singapore University of Technology and Design. She is currently a Post-doctoral Fellow at the Education University of Hong Kong in China. Her research interests mainly focus on computational offloading, edge intelligence, network economics and game theory, and Internet of Things. 
 \end{IEEEbiography}
 
	\begin{IEEEbiography}[{\includegraphics[width=1in,height=1.25in,clip,keepaspectratio]{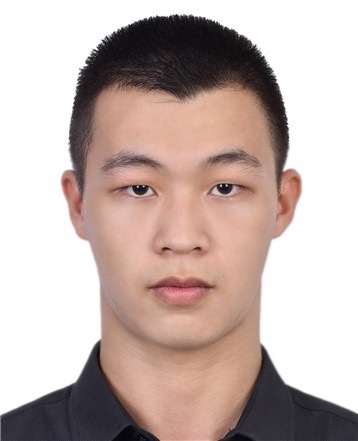}}]{Hui Zeng} received his B.S. degree from Sun Yat-sen University in 2020 and M.S. degree from National University of Defense Technology in 2022. He is currently a Ph.D. student at the National University of Defense Technology. His research interests include data privacy, federated learning, and distributed systems.
\end{IEEEbiography}

	\begin{IEEEbiography}[{\includegraphics[width=1in,height=1.25in,clip,keepaspectratio]{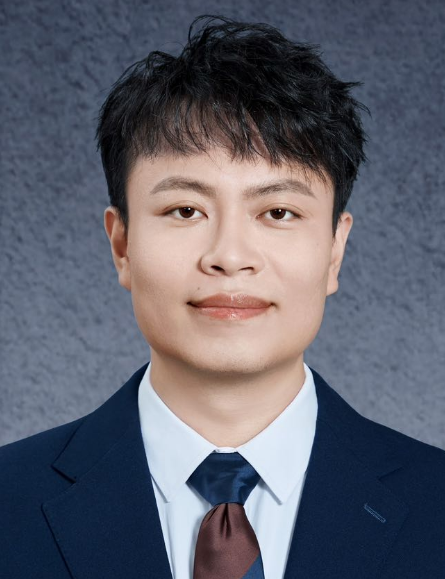}}]{Zehui Xiong} is currently a Full Professor with the School of Electronics, Electrical Engineering and Computer Science, Queen's University Belfast, United Kingdom. Prior to that, he was with Singapore University of Technology and Design, and Alibaba-NTU Singapore Joint Research Institute. He received his Ph.D. degree from Nanyang Technological University and was a visiting scholar with Princeton University and University of Waterloo. Recognized as a Clarivate Highly Cited Researcher, he has published over 250 peer-reviewed research papers in leading journals, with numerous Best Paper Awards from international flagship conferences. His research interests include 5G/B5G/6G and cellular systems, autonomous vehicles and intelligent transportation, cooperative communication and green communication, land transportation, vehicular communications and connected vehicles, wireless networks.
 \end{IEEEbiography}

\begin{IEEEbiography}[{\includegraphics[width=1in,height=1.25in,clip,keepaspectratio]{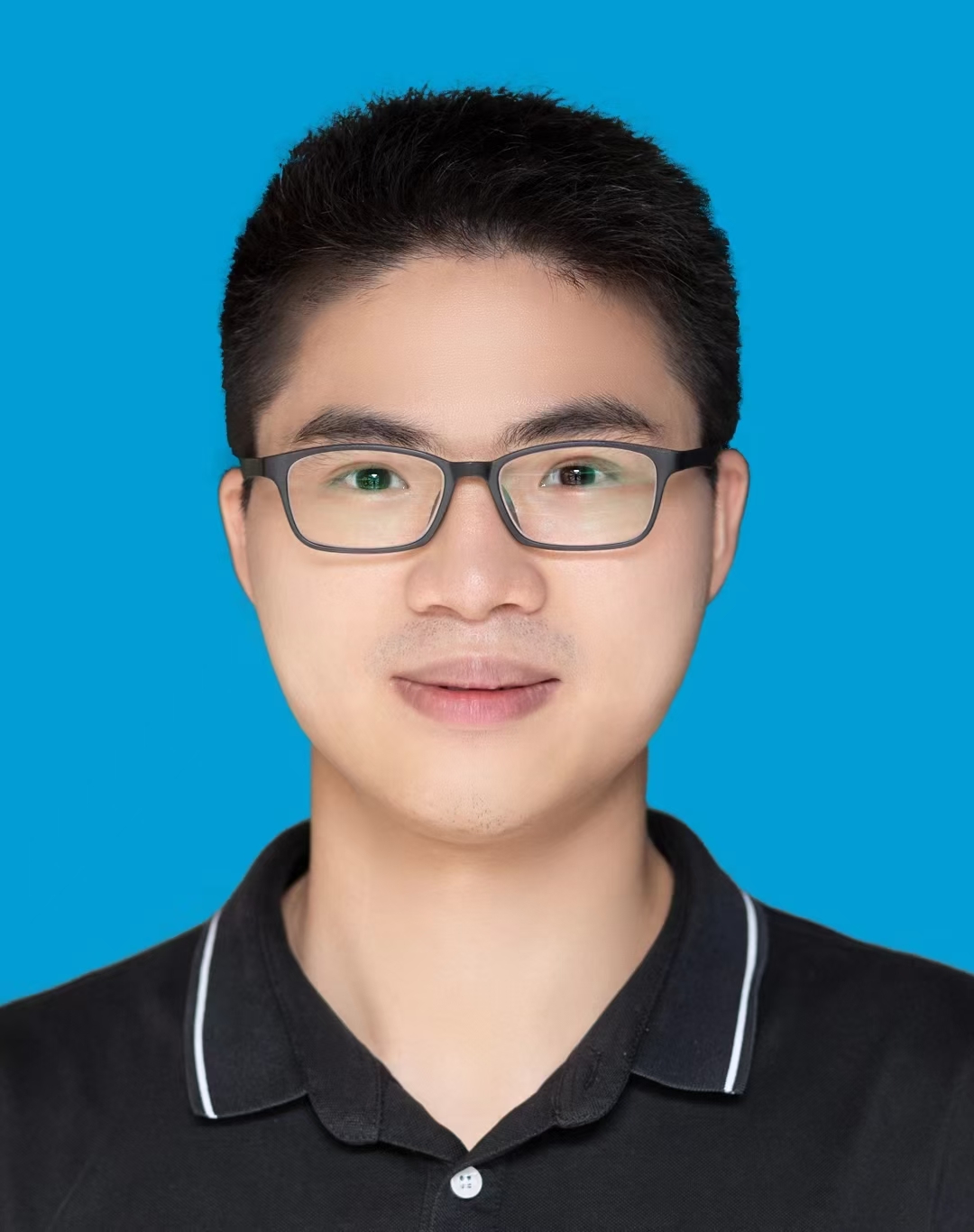}}]{Jiawen Kang} received the M.S. degree and the Ph.D. degree from the Guangdong University of Technology, China, in 2015 and 2018. He is currently a full professor at the Guangdong University of Technology. He has been a postdoc at Nanyang Technological University from 2018 to 2021, Singapore. His research interests mainly focus on blockchain, Metaverse, and AIGC in wireless communications and networking.
\end{IEEEbiography}

\begin{IEEEbiography}[{\includegraphics[width=1in,height=1.25in,clip,keepaspectratio]{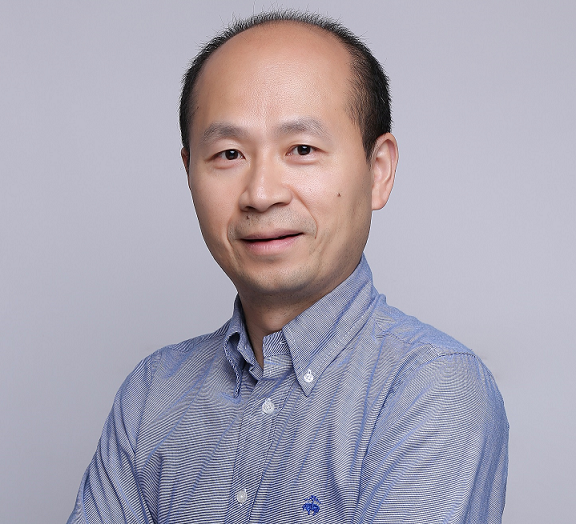}}]{Zhiping Cai}
	received the B.Eng., M.A.Sc., and Ph.D. degrees in computer science and technology from National University of Defense Technology (NUDT), China, in 1996, 2002, and 2005, respectively. He is a full professor in College of Computer Science and Technology, NUDT. His current research interests include artificial intelligence, network security and big data. He is a distinguished member of the CCF.
\end{IEEEbiography}

\begin{IEEEbiography}[{\includegraphics[width=1in,height=1.25in,clip,keepaspectratio]{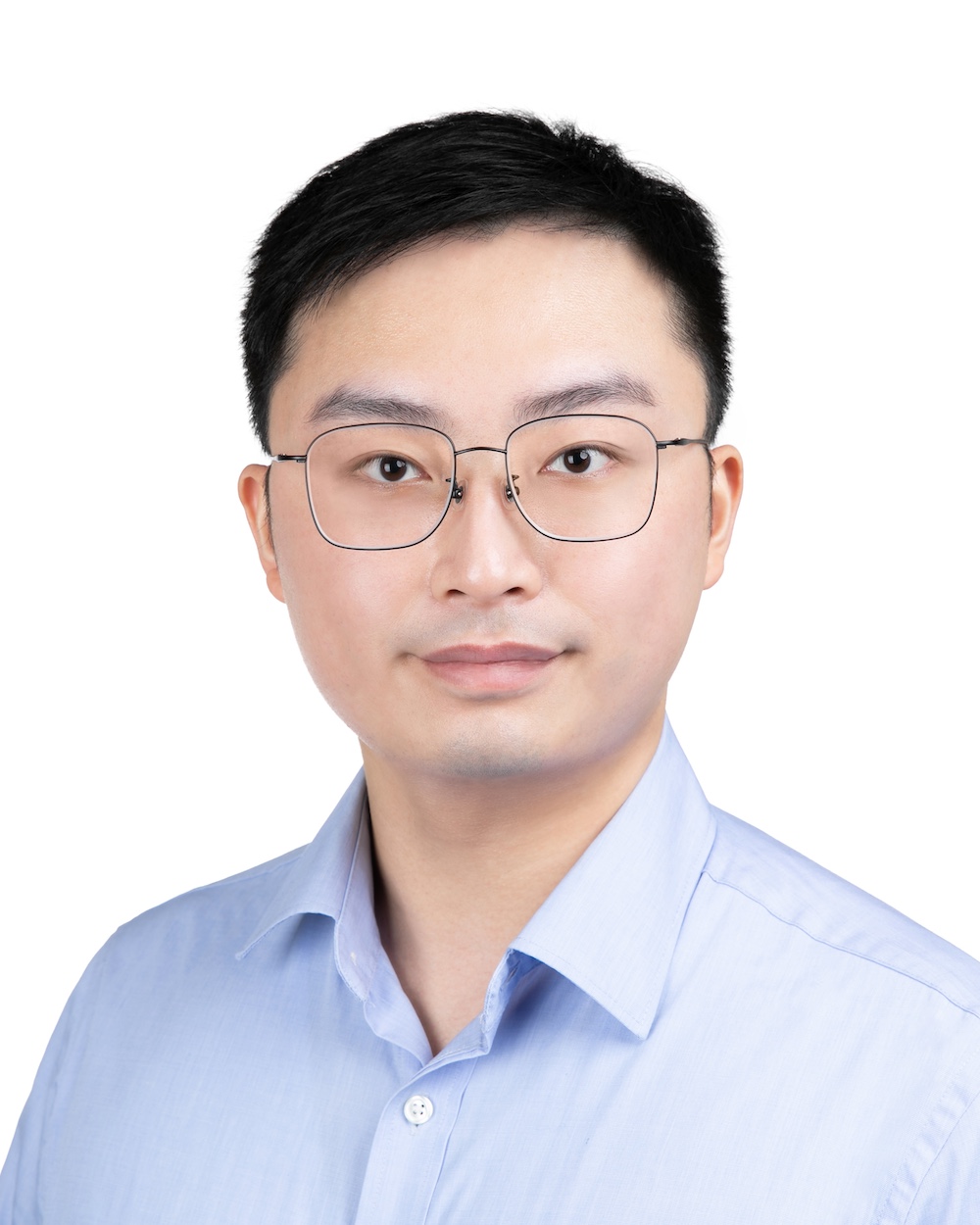}}]{Tse-Tin Chan}(Member, IEEE) received the B.Eng. and Ph.D. degrees in Information Engineering from The Chinese University of Hong Kong (CUHK), Hong Kong, China, in 2014 and 2020, respectively. From 2020 to 2022, he was an Assistant Professor with the Department of Computer Science, The Hang Seng University of Hong Kong (HSUHK), Hong Kong. He is currently an Assistant Professor with the Department of Mathematics and Information Technology, The Education University of Hong Kong (EdUHK), Hong Kong. His research interests include wireless communications and networking, the Internet of Things (IoT), age of information (AoI), and artificial intelligence (AI)-native wireless communications.
\end{IEEEbiography}

\begin{IEEEbiography}[{\includegraphics[width=1in,height=1.25in,clip,keepaspectratio]{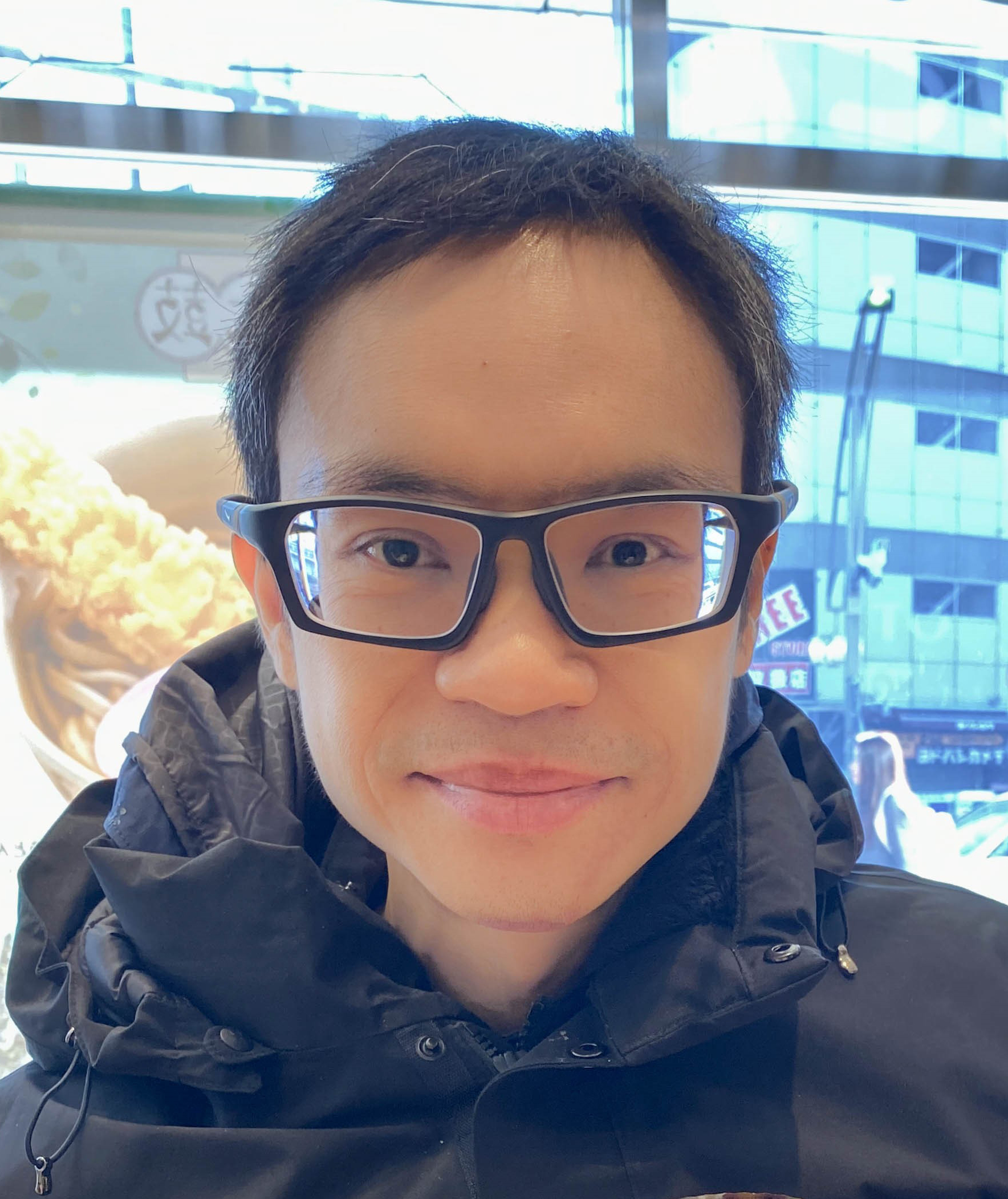}}]{Dusit Niyato}(Fellow, IEEE) is a professor in the School of Computer Science and Engineering, at Nanyang Technological University, Singapore. He received B.Eng. from King Mongkuts Institute of Technology Ladkrabang (KMITL), Thailand in 1999 and Ph.D. in Electrical and Computer Engineering from the University of Manitoba, Canada in 2008. His research interests are in the areas of sustainability, edge intelligence, decentralized machine learning, and incentive mechanism design.
\end{IEEEbiography}

\begin{IEEEbiography}[{\includegraphics[width=1in,height=1.25in,clip,keepaspectratio]{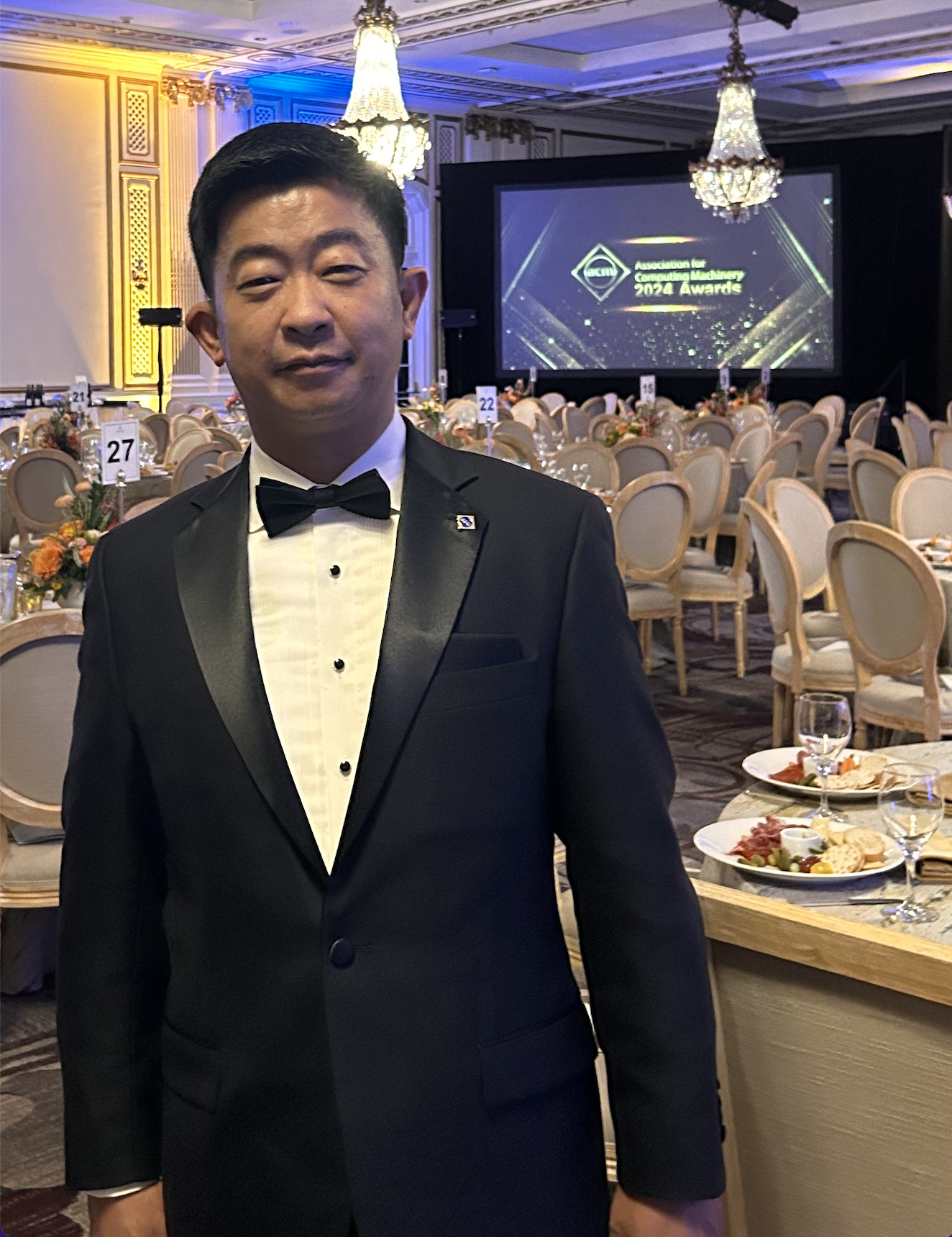}}]{Zhu Han (S'01-M'04-SM'09-F'14)}
	received the B.S. degree in electronic engineering from Tsinghua University, in 1997, and the M.S. and Ph.D. degrees in electrical and computer engineering from the University of Maryland, College Park, in 1999 and 2003, respectively.
	From 2000 to 2002, he was an R\&D Engineer of JDSU, Germantown, Maryland. From 2003 to 2006, he was a Research Associate at the University of Maryland. From 2006 to 2008, he was an assistant professor at Boise State University, Idaho. Currently, he is a John and Rebecca Moores Professor in the Electrical and Computer Engineering Department as well as in the Computer Science Department at the University of Houston, Texas. Dr. Han’s main research targets on the novel game-theory related concepts critical to enabling efficient and distributive use of wireless networks with limited resources. His other research interests include wireless resource allocation and management, wireless communications and networking, quantum computing, data science, smart grid, security and privacy. Dr. Han received an NSF Career Award in 2010, the Fred W. Ellersick Prize of the IEEE Communication Society in 2011, the EURASIP Best Paper Award for the Journal on Advances in Signal Processing in 2015, IEEE Leonard G. Abraham Prize in the field of Communications Systems (best paper award in IEEE JSAC) in 2016, and several best paper awards in IEEE conferences. Dr. Han was an IEEE Communications Society Distinguished Lecturer from 2015-2018, AAAS fellow since 2019, and ACM distinguished Member since 2019. Dr. Han is a 1\% highly cited researcher since 2017 according to Web of Science. Dr. Han is also the winner of the 2021 IEEE Kiyo Tomiyasu Award, for outstanding early to mid-career contributions to technologies holding the promise of innovative applications, with the following citation: ``for contributions to game theory and distributed management of autonomous communication networks."
\end{IEEEbiography}
\end{document}